\documentclass[runningheads]{llncs}

\usepackage{graphicx}
\usepackage{tikz}
\usepackage{comment}
\usepackage{amsmath,amssymb} 
\usepackage{color}

\usepackage{booktabs}
\usepackage[ruled, linesnumbered]{algorithm2e}
\usepackage{enumitem}
\usepackage{multirow}
\usepackage{subcaption}
\usepackage{wrapfig}
\usepackage{scalerel}
\usepackage{xparse}
\usepackage{nicefrac}
\usepackage{xspace}
\usepackage{bbm}
\usepackage{textcomp}
\usepackage{cite}
\usepackage{floatrow}
\usepackage{pifont}
\usepackage{colortbl}
\usepackage{microtype}

\newfloatcommand{capbtabbox}{table}[]
\graphicspath{{figures/}}
\setlist[itemize]{itemsep=0pt,topsep=0pt,partopsep=0pt,parsep=0pt,leftmargin=4mm}
\usepackage[pagebackref,breaklinks,colorlinks]{hyperref}

\usepackage[capitalize]{cleveref}
\crefname{section}{Sec.}{Secs.}
\Crefname{section}{Section}{Sections}
\Crefname{table}{Table}{Tables}
\crefname{table}{Tab.}{Tabs.}

\usepackage[width=122mm,left=12mm,paperwidth=146mm,height=193mm,top=12mm,paperheight=217mm]{geometry}

\makeatletter
\DeclareRobustCommand\onedot{\futurelet\@let@token\@onedot}
\def\@onedot{\ifx\@let@token.\else.\null\fi\xspace}
\def\eg{\emph{e.g}\onedot} 
\def\ie{\emph{i.e}\onedot}

\def\wrt{w.r.t\onedot}

\def\etal{\emph{et al}\onedot}
\makeatother

\newcommand{\red}[1]{\textcolor{red}{#1}}

\definecolor{mygreen}{RGB}{0, 175, 0}
\definecolor{myblue}{RGB}{0, 102, 204}
\definecolor{lightblue}{RGB}{232, 244, 248}
\definecolor{myred}{RGB}{220, 60, 60}
\definecolor{myorange}{RGB}{255, 128, 0}
\newcommand{\myblue}[1]{\textcolor{myblue}{#1}}
\newcommand{\mygreen}[1]{\textcolor{mygreen}{#1}}

\definecolor{hls1}{RGB}{219,  94,  86}
\definecolor{hls2}{RGB}{219, 174,  86}
\definecolor{hls3}{RGB}{184, 219,  86}
\definecolor{hls4}{RGB}{105, 219,  86}
\definecolor{hls5}{RGB}{86, 219, 147}
\definecolor{hls6}{RGB}{86, 211, 219}
\definecolor{hls7}{RGB}{86, 131, 219}
\definecolor{hls8}{RGB}{121,  86, 219}
\definecolor{hls9}{RGB}{200,  86, 219}
\definecolor{hls10}{RGB}{219,  86, 158}
\newcommand{\mv}{\textcolor{hls1}{\textbf{MV}}}
\newcommand{\cam}{\textcolor{hls2}{\textbf{CK}}}
\newcommand{\ck}{\textcolor{hls2}{\textbf{CK}}}
\newcommand{\keypoint}{\textcolor{hls2}{\textbf{CK}}}
\newcommand{\sil}{\textcolor{hls3}{\textbf{S}}}
\newcommand{\pri}{\textcolor{hls4}{\textbf{A}}}
\newcommand{\txt}{\textcolor{hls6}{\textbf{T}}}
\newcommand{\campose}{\textcolor{hls7}{\textbf{P}}}

\newcommand{\dep}{\textcolor{hls9}{\textbf{D}}}
\newcommand{\tm}{$\diamondsuit$\xspace}
\newcommand{\sy}{$\dagger$\xspace}
\newcommand{\rv}{$\ddagger$\xspace}
\newcommand{\nobkg}{$\boxtimes$\xspace}
\newcommand{\notxt}{$\varnothing$\xspace}
\newcommand{\sem}{$\leftrightarrow$\xspace}
\newcommand{\fv}{$\lessdot$\xspace}

\newcommand{\catag}{}
\newcommand{\catagbig}{}
\newcommand{\threed}{\textcolor{myred}{\textbf{3D}\xspace}}
\newcommand{\cmark}{\checkmark}

\newcommand{\multiview}{\textcolor{hls1}{\textbf{M}}ulti-\textcolor{hls1}{\textbf{V}}iews\xspace}

\newcommand{\camerakeypoint}{\textcolor{hls2}{\textbf{C}}amera or 
\textcolor{hls2}{\textbf{K}}eypoints\xspace}
\newcommand{\silhouette}{\textcolor{hls3}{\textbf{S}}ilhouettes\xspace}
\newcommand{\prior}{\textcolor{hls4}{\textbf{A}}ssumptions\xspace}
\newcommand{\camerapose}{\textcolor{hls7}{\textbf{P}}ose\xspace}
\newcommand{\texture}{\textcolor{hls6}{\textbf{T}}exture\xspace}
\newcommand{\ddepth}{\textcolor{hls9}{\textbf{D}}epth\xspace}

\def\RR{{\rm I\!R}}

\def\bA{{\mathbb A}}   \def\bM{{\mathbb M}} \def\bS{{\mathbb S}} 
    \def\bT{{\mathbb T}} 
     
   \def\bP{{\mathbb P}}  
\def\bE{{\mathbb E}}

\def\cA{{\mathcal A}}    \def\cS{{\mathcal S}} 
\def\cB{{\mathcal B}}    \def\cT{{\mathcal T}} 
\def\cC{{\mathcal C}}   \def\cO{{\mathcal O}}

  \def\cL{{\mathcal L}} \def\cR{{\mathcal R}}

\newcommand{\encoder}{e_\theta}
\newcommand{\dmlp}{s_{\theta}}
\newcommand{\tcnn}{t_\theta}
\newcommand{\bcnn}{b_\theta}
\newcommand{\sh}{\textrm{sh}}
\newcommand{\tx}{\textrm{tx}}

\newcommand{\bg}{\textrm{bg}}
\newcommand{\img}{\mathbf{I}}

\newcommand{\rec}{\hat{\mathbf{I}}}
\newcommand{\swapsh}{\hat{\mathbf{I}}_{\textrm{sh}}}
\newcommand{\swaptx}{\hat{\mathbf{I}}_{\textrm{tx}}}
\newcommand{\bkg}{\mathbf{B}}

\newcommand{\pixi}{i}
\newcommand{\facej}{j}
\newcommand{\smesh}{\bS}
\newcommand{\tmesh}{\bT}
\newcommand{\pmesh}{\bP}
\newcommand{\proto}{\bE}
\newcommand{\mesh}{\bM}

\newcommand{\taff}{\cA}
\newcommand{\betaaff}{\mathbf{a}}

\newcommand{\taffbeta}{\taff_{\betaaff}}

\newcommand{\code}{\mathbf{z}}
\newcommand{\betadef}{\mathbf{z}_{\textrm{sh}}}

\newcommand{\tgen}{\cT}
\newcommand{\genbeta}{\mathbf{z}_{\textrm{tx}}}

\newcommand{\bkgbeta}{\mathbf{z}_{\textrm{bg}}}

\newcommand{\rot}{\mathbf{r}}

\newcommand{\scale}{\mathbf{s}}
\newcommand{\trans}{\mathbf{t}}
\newcommand{\prob}{\mathbf{O}}
\newcommand{\col}{\mathbf{C}}

\newcommand{\depth}{\mathbf{D}}
\newcommand{\depthinv}{\depth'}
\newcommand{\dist}{\nu}
\newcommand{\agg}{\cC}
\newcommand{\rend}{\cR}
\newcommand{\aggsr}{\cC_{\textrm{SR}}}
\newcommand{\occ}{\cO}
\newcommand{\occsr}{\cO_{\textrm{SR}}}
\newcommand{\znear}{d_\textrm{near}}
\newcommand{\zfar}{d_\textrm{far}}
\newcommand{\zbkg}{d_\textrm{bg}}

\newcommand{\point}{\mathbf{x}}
\newcommand{\lrec}{\cL_{\textrm{rec}}}
\newcommand{\lcor}{\cL_{\textrm{perc}}}
\newcommand{\lpix}{\cL_{\textrm{pix}}}
\newcommand{\lswap}{\cL_{\textrm{nbr}}}
\newcommand{\lreg}{\cL_{\textrm{reg}}}
\newcommand{\llapl}{\cL_{\textrm{lap}}}

\newcommand{\lnorm}{\cL_{\textrm{norm}}}
\newcommand{\lpose}{\cL_{\textrm{uni}}}
\newcommand{\lmax}{\cL_{\textrm{3D}}}
\newcommand{\lexp}{\cL_{\textrm{P}}}

\newcommand{\wcor}{\lambda_{\textrm{perc}}}

\newcommand{\wswap}{\lambda_{\textrm{nbr}}}
\newcommand{\wreg}{\lambda_{\textrm{reg}}}

\newcommand{\wpose}{\lambda_{\textrm{uni}}}

\newcommand{\probpose}{\mathbf{p}}
\newcommand{\mem}{\Omega}
\newcommand{\midx}{\scaleto{(m)}{6pt}}
\newcommand{\midxs}{\scaleto{(m_{\textrm{s}})}{6pt}}
\newcommand{\midxt}{\scaleto{(m_{\textrm{t}})}{6pt}}

\NewDocumentCommand\curri{}{\scalerel*{\includegraphics{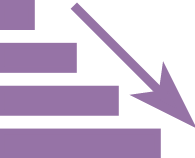}}{X}}

\DeclareMathOperator*{\argmin}{argmin}
\DeclareMathOperator*{\sigmoid}{sigmoid}

\def\httilde{\mbox{\tt\raisebox{.4ex}{\texttildelow}}}

\begin{document}
\pagestyle{headings}
\mainmatter
\def\ECCVSubNumber{1851}  

\title{Share With Thy Neighbors: Single-View Reconstruction by Cross-Instance Consistency} 

\titlerunning{Single-View Reconstruction by Cross-Instance Consistency}

\author{Tom Monnier$^1$ 
  \quad Matthew Fisher$^2$ \quad Alexei A. Efros$^3$ \quad Mathieu Aubry$^1$}
\authorrunning{T. Monnier et al.}
%
\institute{\vspace{-.5em}
$^1$LIGM, Ecole des Ponts, Univ Gustave Eiffel \quad $^2$Adobe Research \quad  $^3$UC 
Berkeley}
\maketitle

\vspace{-.5em}
\begin{figure}[!h]
  \centering
  \captionsetup{width=.93\columnwidth}
  \includegraphics[width=.93\columnwidth]{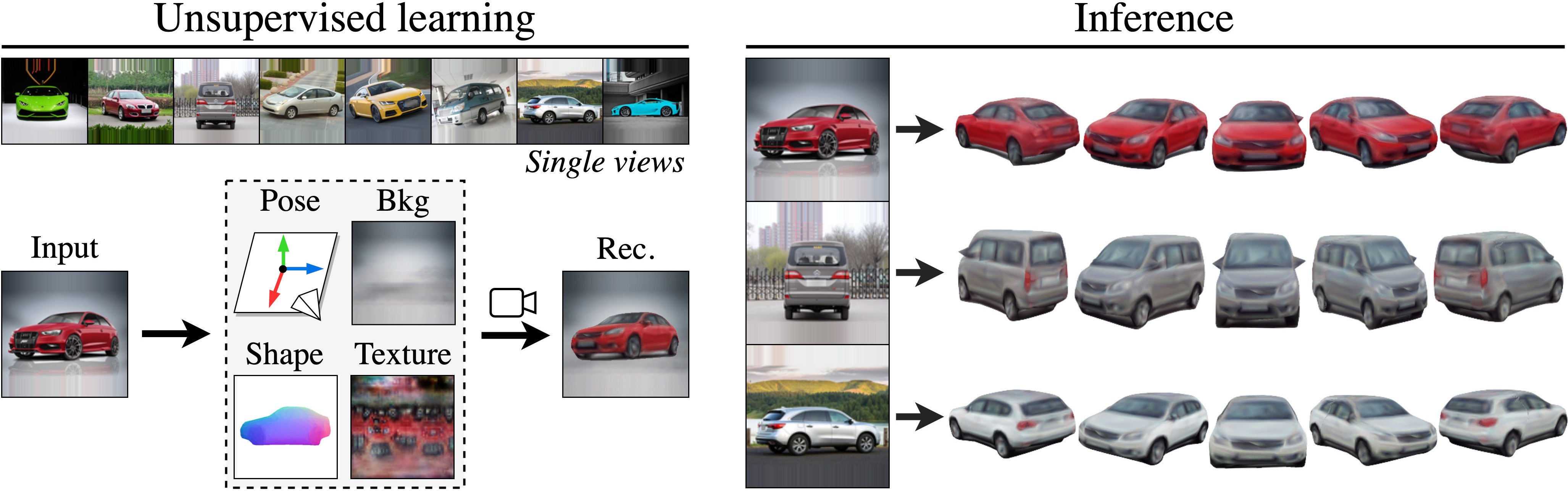}
  \vspace{-.5em}
  \caption{\textbf{Single-View Reconstruction by Cross-Instance Consistency. (left)} Given
    a collection of single-view images from an object category, we learn without additional 
  supervision an autoencoder that explicitly generates shape, texture, pose and background.  
\textbf{(right)} At inference time, our approach reconstructs high-quality textured meshes 
from raw single-view images.}
  \label{fig:teaser}
\end{figure}

\vspace{-1em}
\begin{abstract}
  Approaches for single-view reconstruction typically rely on viewpoint annotations, 
  silhouettes, the absence of background, multiple views of the same instance, a template 
  shape, or symmetry. We avoid all such supervision and assumptions by explicitly
  leveraging the consistency between images of different object instances. As a result, our 
  method can learn from large collections of unlabelled images depicting the same object 
  category. Our main contributions are two ways for leveraging {cross-instance 
  consistency}: (i)~\textit{progressive conditioning}, a training strategy to gradually 
  specialize the model from category to instances in a curriculum learning fashion; and
  (ii)~\textit{neighbor reconstruction}, a loss enforcing consistency between instances 
  having similar shape or texture. Also critical to the success of our method are: our
  structured autoencoding architecture decomposing an image into explicit shape, texture, 
  pose, and background; an adapted formulation of differential rendering; and a new 
  optimization scheme alternating between 3D and pose learning. 
  We compare our approach, UNICORN, both on the diverse synthetic ShapeNet dataset --- the 
  classical benchmark for methods requiring multiple views as supervision --- and on 
  standard real-image benchmarks (Pascal3D+ Car, CUB) for which most methods require known 
  templates and silhouette annotations. We also showcase applicability to more
  challenging real-world collections (CompCars, LSUN), where silhouettes are not available 
  and images are not cropped around the object.


\end{abstract}

\section{Introduction}\label{sec:intro}
\vspace{-.5em}

One of the most magical human perceptual abilities is being able to see the 3D world behind a 
2D image -- a mathematically impossible task!  Indeed, the ancient Greeks were so incredulous 
at the possibility that humans could be ``hallucinating'' the third dimension, that they 
proposed the utterly implausible Emission Theory of Vision~\cite{origins1994} (eye emitting
light to ``sense'' the world) to explain it to themselves.  In the history of computer 
vision, single-view reconstruction (SVR) has had an almost cult status as one of the holy 
grail problems~\cite{hoiem2005geometric, hoiem2008putting, saxenaMake3DLearning3D2009}.  
Recent advancements in deep learning methods have dramatically improved results in this 
area~\cite{choy3DR2N2UnifiedApproach2016, meschederOccupancyNetworksLearning2019}. However, 
the best methods still require costly supervision at training time, such as multiple 
views~\cite{liuSoftRasterizerDifferentiable2019, 
niemeyerDifferentiableVolumetricRendering2020}. Despite efforts to remove such requirements, 
the works with the least supervision still rely on two signals limiting their applicability: 
(i) silhouettes and (ii) strong assumptions such as 
symmetries~\cite{kanazawaLearningCategorySpecificMesh2018, huSelfSupervised3DMesh2021}, known 
template shapes~\cite{goelShapeViewpointKeypoints2020, 
tulsianiImplicitMeshReconstruction2020}, or the absence of 
background~\cite{wuUnsupervisedLearningProbably2020}. Although crucial to achieve reasonable 
results, priors like silhouettes and symmetry can also harm the reconstruction quality:
silhouette annotations are often coarse~\cite{chenLearningPredict3D2019} and small symmetry 
prediction errors can yield unrealistic 
reconstructions~\cite{wuUnsupervisedLearningProbably2020, goelShapeViewpointKeypoints2020}. 

\begin{table}[!b]
  \vspace{-.5em}
  \renewcommand{\arraystretch}{0.95}
  \addtolength{\tabcolsep}{1pt}
  \centering
  \scriptsize
  \begin{tabular}{@{}llll@{}} \toprule
    Method & Supervision & Data & Output\\
    \midrule

    Pix2Mesh~\cite{wangPixel2MeshGenerating3D2018},
    AtlasNet~\cite{groueixAtlasNetPapierMAch2018},
    OccNet~\cite{meschederOccupancyNetworksLearning2019}\catag
  & \threed & ShapeNet & 3D\catagbig\\
  
  PTN\catag~\cite{yanPerspectiveTransformerNets2016}, NMR\catag~\cite{katoNeural3DMesh2018}
  & \mv, \cam, \sil  & ShapeNet & 3D\catagbig \\

  DRC\catag~\cite{tulsianiMultiviewSupervisionSingleview2017},
  SoftRas\catag~\cite{liuSoftRasterizerDifferentiable2019}, 
  DVR\catag~\cite{niemeyerDifferentiableVolumetricRendering2020}
  & \mv, \cam, \sil & ShapeNet & 3D, \txt \catagbig\\
  
  GANverse3D~\cite{zhangImageGANsMeet2021} & \mv, \cam, \sil & Bird, Car, Horse& 3D, \txt\\

  DPC\catag~\cite{insafutdinovUnsupervisedLearningShape2018},
  MVC\catag~\cite{tulsianiMultiviewConsistencySupervisory2018}
  & \mv, \sil & ShapeNet & 3D, \campose\catagbig\\

  Vicente \etal~\cite{vicenteReconstructingPASCALVOC2014},
  CSDM~\cite{ karCategorySpecificObjectReconstruction2015},
  DRC~\cite{tulsianiMultiviewSupervisionSingleview2017}\catag
  & \cam, \sil & Pascal3D & 3D\catagbig\\

  CMR~\cite{kanazawaLearningCategorySpecificMesh2018}
  & \keypoint, \sil, \pri (\sy) & Bird, Car, Plane &3D, \txt\\

  SDF-SRN\cite{linSDFSRNLearningSigned2020},
  TARS~\cite{duggalTopologicallyAwareDeformationFields2022} & \cam, \sil & ShapeNet, Bird, 
  Car, Plane & 3D, \txt\\

  TexturedMeshGen\cite{hendersonLeveraging2DData2020}
  & \cam, \pri (\sy) & ShapeNet, Bird, Car & 3D, \txt\\

  UCMR\cite{goelShapeViewpointKeypoints2020},
  IMR~\cite{tulsianiImplicitMeshReconstruction2020}
  & \sil, \pri (\tm, \sy) & Animal, Car, Moto & 3D, \txt, \campose\\

  UMR~\cite{liSelfsupervisedSingleview3D2020}
  & \sil, \pri (\sem, \sy) & Animal, Car, Moto & 3D, \txt, \campose\\

  RADAR~\cite{wuDerenderingWorldRevolutionary2021}
  & \sil, \pri (\rv) & Vase & 3D, \txt, \campose\\

  SMR~\cite{huSelfSupervised3DMesh2021} & \sil, \pri (\sy) & ShapeNet, Animal, Moto & 3D, 
  \txt, \campose\\

  Unsup3D~\cite{wuUnsupervisedLearningProbably2020}
  & \pri (\nobkg, \fv, \sy) & Face & \dep, \txt, \campose\\

  Henderson \& Ferrari~\cite{hendersonLearningSingleimage3D2019}
  & \pri (\nobkg, \notxt) & ShapeNet & 3D, \campose\\

  \textbf{Ours} & None & ShapeNet, Animal, Car, Moto & 3D, \txt, \campose \\

  \bottomrule
  \end{tabular}
  \vspace{-.8em}
  \caption{\textbf{Comparison with selected works.} For each method, we outline the 
    supervision and priors used (\threed, \multiview, \camerakeypoint, \silhouette, \prior 
      like \tm template shape, \sy symmetry, \rv solid of revolution, \sem semantic 
      consistency, \nobkg no/limited background,
  \fv frontal view, \notxt no texture), which data it has been applied to and the model 
  output (3D, \texture, \camerapose, \ddepth).
}

  \label{tab:comp}
\end{table}

\begin{figure}[!t]
  \centering
  \begin{subfigure}[t]{0.38\textwidth}
    \centering
    \includegraphics[width=\columnwidth]{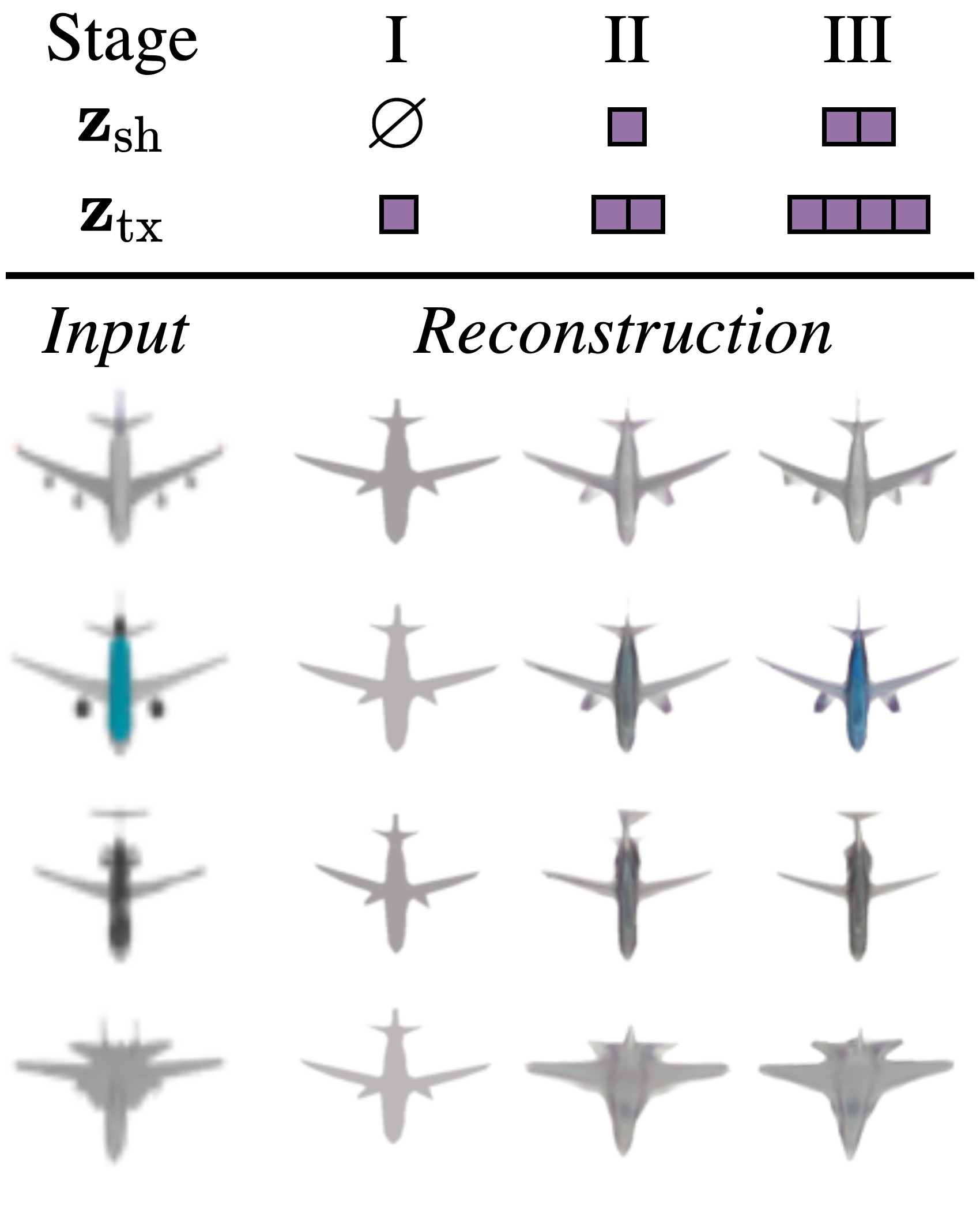}
    \caption{Progressive conditioning}
    \label{fig:teaser_curri}
  \end{subfigure}\hspace*{\fill}
  \begin{subfigure}[t]{0.555\textwidth}
    \includegraphics[width=\columnwidth]{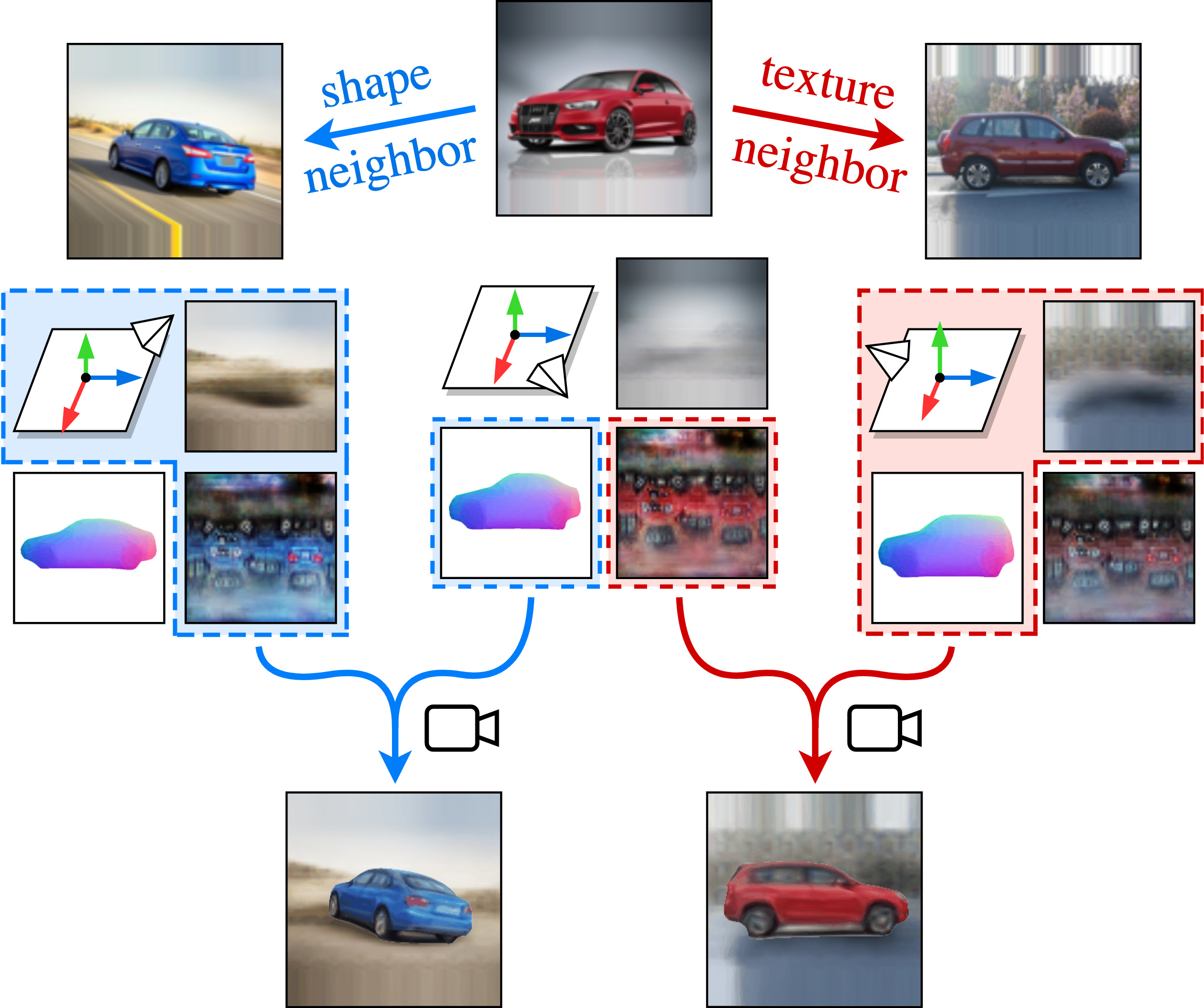}
    \caption{Neighbor reconstruction}
    \label{fig:teaser_swap}
  \end{subfigure}
  \vspace{-.8em}
  \caption{\textbf{Leveraging cross-instance consistency. (a)} Progressive conditioning 
  amounts to gradually increasing, in a multi-stage fashion, the size of the conditioning
  latent spaces, here associated to shape $\code_\sh$ and texture $\code_\tx$. \textbf{(b)}
  We explicitly share the shape and texture models across neighboring instances by swapping 
their characteristics and applying a loss to associated neighbor 
reconstructions.\vspace{-.7em}}
    \label{fig:unicorn}
\end{figure}

In this paper, we propose the most unsupervised approach to single-view reconstruction to 
date, which we demonstrate to be competitive for diverse datasets. \Cref{tab:comp} summarizes 
the differences between our approach and representative prior works.
More precisely, we learn in an analysis-by-synthesis fashion a network that predicts 
for each input image: 1) a 3D shape parametrized as a deformation of an ellipsoid, 2) a
texture map, 3) a camera viewpoint, and 4) a background image (\cref{fig:teaser}). Our main 
insight to remove the supervision and assumptions required by other methods is to leverage 
the consistency across different instances.  First, we design a training procedure, {\it 
progressive conditioning}, which encourages the model to share elements between images by 
strongly constraining the variability of shape, texture and background at the beginning of 
training and progressively allowing for more diversity~(\cref{fig:teaser_curri}). Second, we 
introduce a {\it neighbor reconstruction} loss, which explicitly enforces neighboring 
instances from different viewpoints to share the same shape or texture 
model~(\cref{fig:teaser_swap}). Note that these simple yet effective techniques are 
data-driven and not specific to any dataset.  Our only remaining assumption is the knowledge 
of the semantic class of the depicted object. 

We also provide two technical insights that we found critical to learn our model without 
viewpoint and silhouette annotations:
(i) a differentiable rendering formulation inspired by layered image 
models~\cite{jojicLearningFlexibleSprites2001, monnierUnsupervisedLayeredImage2021} which we 
found to perform better than the classical
SoftRasterizer~\cite{liuSoftRasterizerDifferentiable2019}, and (ii) a new optimization 
strategy which alternates between learning a set of pose candidates with associated 
probabilities and learning all other components using the most likely candidate.

We validate our approach on the standard ShapeNet~\cite{changShapeNetInformationRich3D2015} 
benchmark, real image SVR benchmarks
(Pascal3D+ Car~\cite{xiangPASCALBenchmark3D2014}, CUB~\cite{welinderCaltechucsdBirds2002010}) 
as well as more complex real-world datasets (CompCars~\cite{yangLargescaleCarDataset2015}, 
LSUN Motorbike and Horse~\cite{yuLSUNConstructionLargescale2016}). In all scenarios, we 
demonstrate results competitive with the best supervised methods.

\vspace{-1em}
\subsubsection{Summary.}  We present UNICORN, a framework leveraging \textbf{UN}supervised 
cross-\textbf{I}nstance \textbf{CO}nsistency for 3D \textbf{R}econstructio\textbf{N}. Our 
main contributions are: 1) the most unsupervised SVR system to date, demonstrating 
state-of-the-art textured 3D reconstructions for both generic shapes and real images, and not 
requiring supervision or restrictive assumptions beyond a categorical image collection; 2) 
two data-driven techniques to enforce cross-instance consistency, namely progressive 
conditioning and neighbor reconstruction. Code and video results are available at 
\href{http://imagine.enpc.fr/~monniert/UNICORN/}{\tt imagine.enpc.fr/\httilde 
monniert/UNICORN}.

\section{Related work}\label{sec:related}
\vspace{-.5em}

We first review deep SVR methods and mesh-based differential renderers we build upon. We then 
discuss works exploring cross-instance consistency and curriculum learning techniques, to 
which our progressive conditioning is related.

\vspace{-1.2em}
\subsubsection{Deep SVR.} There is a clear trend to remove supervision from deep SVR 
pipelines to directly learn 3D from raw 2D images, which we summarize in~\Cref{tab:comp}.

A first group of methods uses strong supervision, either paired 3D and images or multiple 
views of the same object. Direct 3D supervision is successfully used to learn 
voxels~\cite{choy3DR2N2UnifiedApproach2016},
meshes~\cite{wangPixel2MeshGenerating3D2018},
parametrized surfaces~\cite{groueixAtlasNetPapierMAch2018} and implicit 
functions~\cite{meschederOccupancyNetworksLearning2019, xuDISNDeepImplicit2019}. The first 
methods using silhouettes and multiple views initially require camera poses and are also 
developed for diverse 3D shape representations: \cite{yanPerspectiveTransformerNets2016, 
tulsianiMultiviewSupervisionSingleview2017} opt for voxels, \cite{katoNeural3DMesh2018, 
liuSoftRasterizerDifferentiable2019, chenLearningPredict3D2019} introduce mesh renderers, 
and~\cite{niemeyerDifferentiableVolumetricRendering2020} adapts implicit representations.  
Works like~\cite{insafutdinovUnsupervisedLearningShape2018, 
tulsianiMultiviewConsistencySupervisory2018} then introduce techniques to remove the 
assumption of known poses. Except for~\cite{zhangImageGANsMeet2021} which leverages 
GAN-generated images~\cite{goodfellowGenerativeAdversarialNets2014, 
karrasStyleBasedGeneratorArchitecture2019}, these works are typically limited to synthetic 
datasets.

A second group of methods aims at removing the need for 3D and multi-view supervision. This 
is very challenging and they hence typically focus on learning 3D from images of a single 
category.  Early works~\cite{vicenteReconstructingPASCALVOC2014, 
karCategorySpecificObjectReconstruction2015, tulsianiMultiviewSupervisionSingleview2017} 
estimate camera poses with keypoints and minimize the silhouette reprojection error. The 
ability to predict textures is first incorporated by 
CMR~\cite{kanazawaLearningCategorySpecificMesh2018} which, in addition to keypoints and 
silhouettes, uses symmetry priors. Recent works~\cite{linSDFSRNLearningSigned2020, 
duggalTopologicallyAwareDeformationFields2022} replace the mesh representation of CMR with 
implicit functions that do not require symmetry priors, yet the predicted texture quality is 
strongly deteriorated. \cite{hendersonLeveraging2DData2020} improves upon CMR and develops a 
framework for images with camera annotations that does not rely on silhouettes. Two works 
managed to further avoid the need for camera estimates but at the cost of additional 
hypothesis:~\cite{hendersonLearningSingleimage3D2019} shows results with textureless 
synthetic objects,~\cite{wuUnsupervisedLearningProbably2020} models 2.5D objects like faces 
with limited background and viewpoint variation. Finally, recent works only require object 
silhouettes but they also make additional assumptions:~\cite{goelShapeViewpointKeypoints2020, 
tulsianiImplicitMeshReconstruction2020} use known template 
shapes,~\cite{liSelfsupervisedSingleview3D2020} assumes access to an off-the-shelf system 
predicting part semantics, and~\cite{wuDerenderingWorldRevolutionary2021} targets solids of 
revolution. Other related works~\cite{gadelha3DShapeInduction2017, 
  katoLearningViewPriors2019, henzlerEscapingPlatoCave2019, 
pavlloConvolutionalGenerationTextured2020, yeShelfSupervisedMeshPrediction2021, 
huSelfSupervised3DMesh2021} leverage in addition generative adversarial techniques to improve 
the learning.

In this work, we \textit{do not} use camera estimates,
keypoints, silhouettes, nor strong dataset-specific assumptions, and demonstrate results for 
both diverse shapes and real images. To the best of our knowledge, we present the first 
generic SVR system learned from raw image collections.

\vspace{-1.2em}
\subsubsection{Mesh-based differentiable rendering.} We represent 3D models as meshes with 
parametrized surfaces, as introduced in AtlasNet~\cite{groueixAtlasNetPapierMAch2018} and 
advocated by~\cite{tulsianiImplicitMeshReconstruction2020}. We optimize the mesh geometry, 
texture and camera parameters associated to an image using differentiable rendering. Loper 
and Black~\cite{loperOpenDRApproximateDifferentiable2014} introduce the first generic 
differentiable renderer by approximating derivatives with local filters, 
and~\cite{katoNeural3DMesh2018} proposes an alternative approximation more suitable to 
learning neural networks. Another set of methods instead approximates the rendering function 
to allow differentiability, including 
SoftRasterizer~\cite{liuSoftRasterizerDifferentiable2019,raviAccelerating3DDeep2020} and 
DIB-R~\cite{chenLearningPredict3D2019}. We refer the reader 
to~\cite{katoDifferentiableRenderingSurvey2020} for a comprehensive study. We build upon
SoftRasterizer~\cite{liuSoftRasterizerDifferentiable2019}, but modify the rendering function 
to learn without silhouette information.

\vspace{-1.2em}
\subsubsection{Cross-instance consistency.} Although all methods learned on categorical image 
collections implicitly leverage the consistency across instances, few recent works explicitly 
explore such a signal.
Inspired by~\cite{kulkarniCanonicalSurfaceMapping2019}, the SVR system 
of~\cite{huSelfSupervised3DMesh2021} is learned by enforcing consistency between the 
interpolated 3D attributes of two instances and attributes predicted for the associated 
reconstruction. \cite{yaoDiscovering3DParts2021} discovers 3D parts using the inconsistency 
of parts across
instances. Closer to our approach,~\cite{navaneetImageCollectionsPoint2020} introduces a loss 
enforcing cross-silhouette consistency. Yet it differs from our work in two ways: (i) the 
loss operates on silhouettes, whereas our loss is adapted to image reconstruction by modeling 
background and separating two terms related to shape and texture, and (ii) the loss is used 
as a refinement on top of two cycle consistency losses for poses and 3D reconstructions, 
whereas we demonstrate results without additional self-supervised losses. 

\vspace{-1.2em}
\subsubsection{Curriculum learning.}
The idea of learning networks by ``starting small'' dates back to 
Elman~\cite{elmanLearningDevelopmentNeural1993} where two curriculum learning schemes are 
studied: (i) increasing the difficulty of samples, and (ii) increasing the model complexity.  
We respectively coin them \textit{curriculum sampling} and \textit{curriculum modeling} for 
differentiation. Known to drastically improve the convergence 
speed~\cite{bengioCurriculumLearning2009}, curriculum sampling is widely adopted across 
various applications~\cite{schroffFaceNetUnifiedEmbedding2015, 
bengioScheduledSamplingSequence2015, ilgFlowNetEvolutionOptical2017}. On the contrary,
curriculum modeling is typically less studied although crucial to various methods. For 
example,~\cite{wangPixel2MeshGenerating3D2018} performs SVR in a coarse-to-fine manner by 
increasing the number of mesh vertices, 
and~\cite{monnierDeepTransformationInvariantClustering2020} clusters images by aligning them 
with transformations that increase in complexity.
We propose a new form of curriculum modeling dubbed \textit{progressive conditioning} which 
enables us to avoid bad minima.
\vspace{-1.3em}

\section{Approach}\label{sec:method}
\vspace{-.2em}

\begin{figure}[t]
    \centering
    \includegraphics[width=\columnwidth]{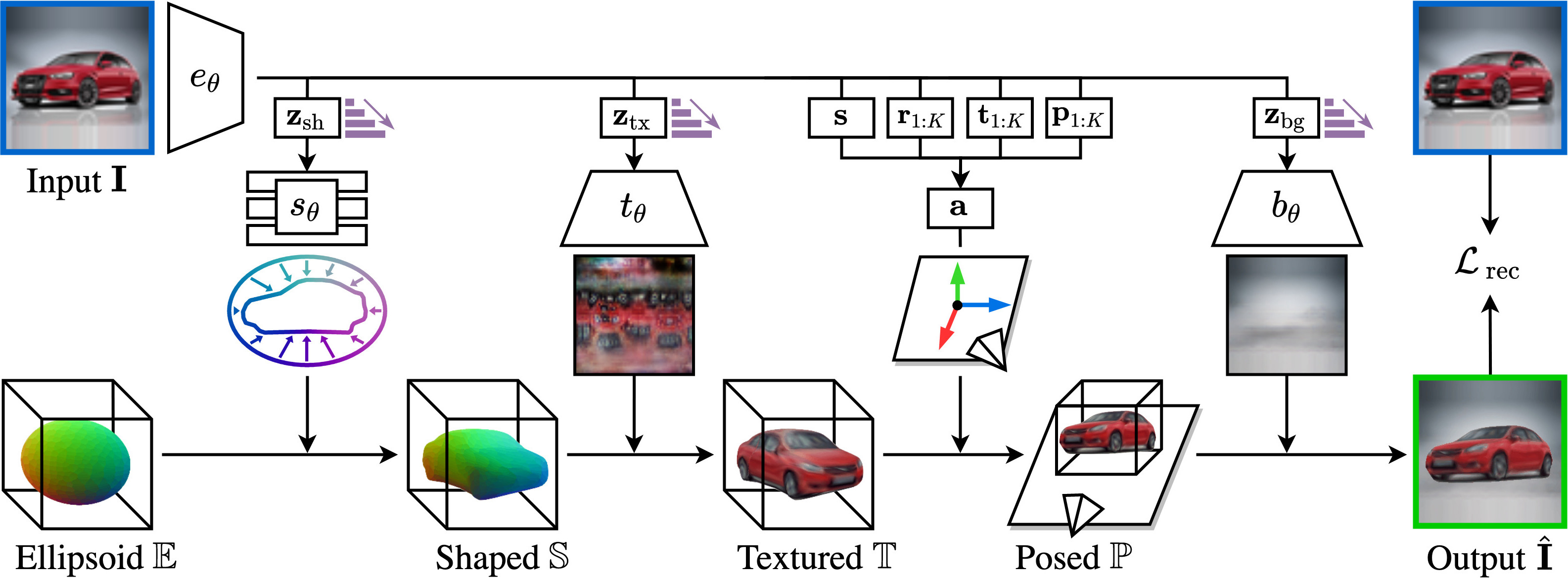}
    \vspace{-1.3em}
    \caption{\textbf{Structured autoencoding.} Given an \myblue{input}, we 
      predict parameters that are decoded into 4 factors (shape, texture, pose, background) 
      and composed to generate the \mygreen{output}. Progressive conditioning is represented 
      with \curri.  \vspace{-.3em}
  }
  \label{fig:pipeline}
\end{figure}

Our goal is to learn a neural network that reconstructs a textured 3D object from a single 
input image. We assume we have access to a raw collection of images depicting objects from 
the same category, without any further annotation. We propose to learn in an 
analysis-by-synthesis fashion by autoencoding images in a structured way 
(\cref{fig:pipeline}). We first introduce our structured autoencoder (\cref{sec:obj}). We 
then present how we learn models consistent across instances (\cref{sec:learning}). Finally, 
we discuss one more technical contribution necessary to our system: an alternate optimization 
strategy for joint 3D and pose estimation (\cref{sec:optim}).

\vspace{-1.2em}
\subsubsection{Notations.} We use bold lowercase for vectors (e.g., $\mathbf{a}$), bold 
uppercase for images (e.g., $\mathbf{A}$), double-struck uppercase for meshes (e.g., $\bA$), 
calligraphic uppercase for the main modules of our system (e.g., $\mathcal{A}$), lowercase 
indexed with generic parameters 
$\theta$ for networks (e.g., $a_\theta$), and write 
$a_{1:N}$ the ordered set $\{a_1,\ldots,a_n\}$.

\vspace{-.5em}
\subsection{Structured autoencoding}\label{sec:obj}

\subsubsection{Overview.} Our approach can be seen as a structured autoencoder: it takes an 
image as input, computes parameters with an encoder, and decodes them into explicit and 
interpretable factors that are composed to generate an image. We model images as the 
rendering of textured meshes on top of background images. For a given image $\img$, our model 
thus predicts a shape, a texture, a pose and a background which are composed to get the 
reconstruction $\rec$, as shown in~\Cref{fig:pipeline}. More specifically, the image $\img$ 
is fed to convolutional encoder networks $\encoder$ which output parameters $\encoder(\img) = 
\{\code_\sh,\code_\tx,\betaaff,\code_\bg\}$ used for the decoding part.  $\betaaff$ is a 9D 
vector including the object pose, while the dimension of the latent codes 
$\code_\sh,\code_\tx$ and $\code_\bg$ will vary during training (see \cref{sec:learning}).  
In the following, we describe the decoding modules using these parameters to build the final 
image by generating a shape, adding texture, positioning it and rendering it over a 
background.

\vspace{-1em}
\subsubsection{Shape deformation.} We follow~\cite{tulsianiImplicitMeshReconstruction2020} 
and use the parametrization of AtlasNet~\cite{groueixAtlasNetPapierMAch2018} where different 
shapes are represented as
deformation fields applied to the unit sphere. We apply the deformation to 
an icosphere slightly stretched into 
an ellipsoid mesh $\proto$ using a fixed anisotropic scaling. 
More specifically, given a 3D vertex $\point$ of the ellipsoid, 
our shape deformation module $\cS_{\code_\sh}$ is defined by $\cS_{\code_\sh}(\point) = 
\point + \dmlp(\point, \code_\sh),$ where $\dmlp$ is a Multi-Layer Perceptron taking as input 
the concatenation of a 3D point $\point$ and a shape code $\code_\sh$. Applying this 
displacement to all the ellipsoid vertices enables us to generate a shaped mesh $\smesh = 
\cS_{\code_\sh}(\proto)$.  We found that using an ellipsoid instead of a raw icosphere was 
very effective in encouraging the learning of objects aligned \wrt the canonical axes.  
Learning surface deformations is often preferred to vertex-wise displacements as it enables 
mapping surfaces, and thus meshes, at any resolution.  For us, the mesh resolution is kept 
fixed and such a representation is a way to regularize the deformations.

\vspace{-1em}
\subsubsection{Texturing.} Following the idea of 
CMR~\cite{kanazawaLearningCategorySpecificMesh2018}, we model textures as an image UV-mapped 
onto the mesh through the reference ellipsoid. More specifically, given a texture code 
$\code_\tx$, a convolutional network $\tcnn$ is used to produce an image $\tcnn(\code_\tx)$,  
which is UV-mapped onto the sphere using spherical coordinates to associate a 2D point to 
every vertex of the ellipsoid, and thus to each vertex of the shaped mesh. We write 
$\tgen_{\code_\tx}$ this module generating a textured mesh $\tmesh = 
\tgen_{\code_\tx}(\smesh)$.

\vspace{-1em}
\subsubsection{Affine transformation.} To render the textured mesh $\tmesh$, we define its 
position \wrt the camera. In addition, we found it beneficial to explicitly model an 
anisotropic scaling of the objects. Because predicting poses is difficult, we predict $K$ 
poses candidates, defined by rotations $\rot_{1:K}$ and
translations $\trans_{1:K}$, and associated probabilities $\probpose_{1:K}$. This involves 
learning challenges we tackle with a specific optimization procedure described 
in~\cref{sec:optim}. At inference, we select the pose with highest probability. We combine 
the scaling and the most likely 6D pose in a single affine transformation module $\taffbeta$.  
More precisely, $\taffbeta$ is parametrized by $\betaaff = \{\scale, \rot, \trans\}$, where 
$\scale, \rot, \trans \in \RR^3$ respectively correspond to the three scales of an 
anisotropic scaling, the three Euler angles of a rotation and the three coordinates of a 
translation. A 3D point $\point$ on the mesh is then transformed by $\taffbeta(\point) = 
\text{rot}(\rot)\text{diag}(\scale)\point + \trans$ where $\text{rot}(\rot)$ is the rotation 
matrix associated to $\rot$ and $\text{diag}(\scale)$ is the diagonal matrix associated to 
$\scale$. Our module is applied to all points of the textured mesh $\tmesh$ resulting in a 
posed mesh $\pmesh=\taff_{\betaaff}(\tmesh)$.

\vspace{-1em}
\subsubsection{Rendering with background.} The final step of our process is to render the 
mesh over a background image. The background image is generated from a background code 
$\code_\bg$ by a convolutional network $\bcnn$.  A differentiable module $\cB_{\code_\bg}$ 
renders the posed mesh $\pmesh$ over this background image $\bcnn(\code_\bg)$ resulting in a 
reconstructed image $\rec = \cB_{\code_\bg}(\pmesh)$.  We perform rendering through soft 
rasterization of the mesh. Because we observed divergence results when learning geometry from 
raw photometric comparison with the standard 
SoftRasterizer~\cite{liuSoftRasterizerDifferentiable2019, raviAccelerating3DDeep2020}, we 
propose two key changes: a layered aggregation of the projected faces and an alternative 
occupancy function. We provide details in our supplementary material.

\subsection{Unsupervised learning with cross-instance consistency}\label{sec:learning}

\begin{figure}[t]
  \centering
  \begin{subfigure}[t]{0.46\textwidth}
    \includegraphics[width=\columnwidth]{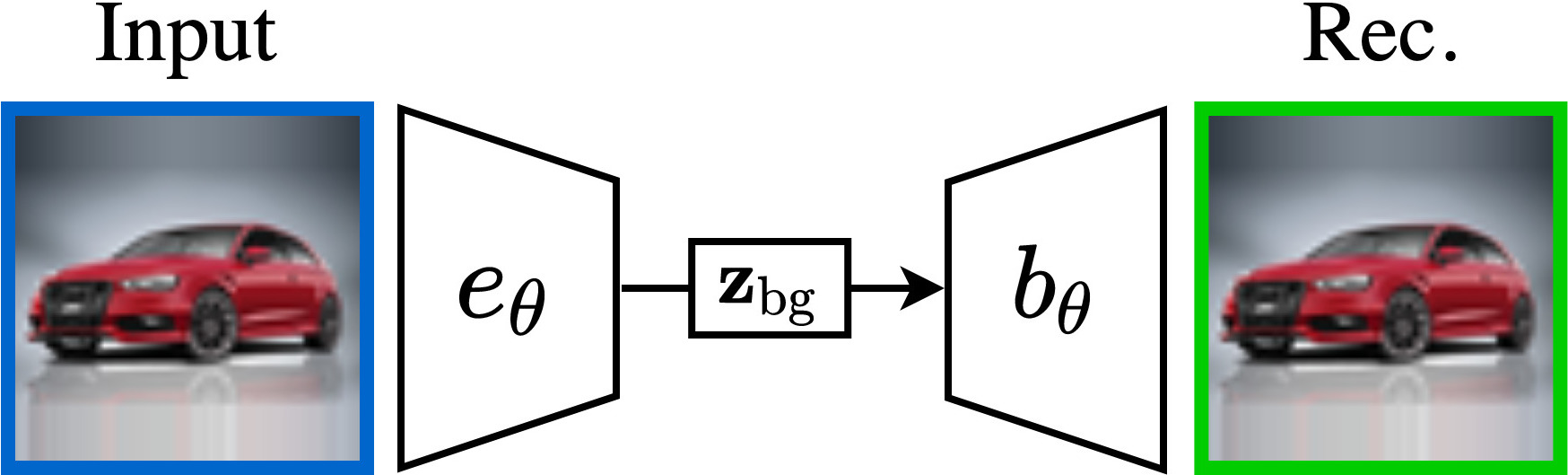}
    \caption{Degenerate background}
    \label{fig:overfit_bg}
  \end{subfigure}\hspace*{\fill}
  \begin{subfigure}[t]{0.48\textwidth}
    \centering
    \includegraphics[width=\columnwidth]{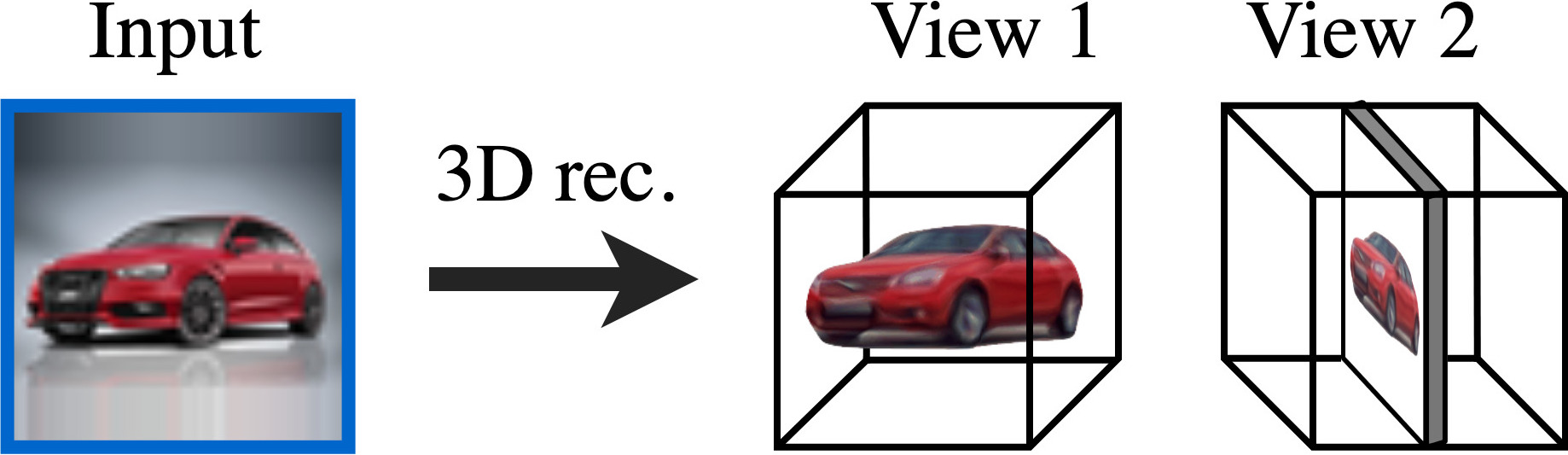}
    \caption{Degenerate 3D object}
  \label{fig:overfit_3d}
  \end{subfigure}
  \caption{\textbf{Degenerate solutions.} An SVR system learned by raw image autoencoding is 
  prone to degenerate solutions through \textbf{(a)} the background or \textbf{(b)} the 3D 
object model.  We alleviate the issue with cross-instance consistency.  \vspace{-.5em}}
   \vspace{-.5em}
\end{figure}

We propose to learn our structured autoencoder without any supervision, by synthesizing 2D 
images and minimizing a reconstruction loss. Due to the unconstrained nature of the problem, 
such an approach typically yields degenerate solutions (\cref{fig:overfit_bg}
and~\cref{fig:overfit_3d}). While previous works leverage silhouettes and dataset-specific 
priors to mitigate this issue, we instead propose two unsupervised data-driven techniques, 
namely
{\it progressive conditioning} (a training strategy) and {\it neighbor reconstruction} (a 
training loss). We thus optimize the shape, texture and background by minimizing for each
image $\img$ reconstructed as $\rec$:%
\begin{equation}\label{eq:lmax}
  \lmax = \lrec(\img, \rec) + \wswap\lswap + \wreg\lreg,
\end{equation}%
where $\wswap$ and $\wreg$ are scalar hyperparameters, and $\lrec$, $\lswap$ and $\lreg$ are
respectively the reconstruction, neighbor reconstruction, and regularization losses, 
described below.  In all experiments, we use $\wswap = 1$ and  $\wreg = 0.01$. Note that we 
optimize pose prediction using a slightly different loss in an alternate optimization scheme 
described in~\Cref{sec:optim}. 

\vspace{-1em}
\subsubsection{Reconstruction and regularization losses.} Our reconstruction loss has two 
terms, a pixel-wise squared $L_2$ loss $\lpix$ and a perceptual 
loss~\cite{zhangUnreasonableEffectivenessDeep2018} $\lcor$ defined as an $L_2$ loss on the 
\texttt{relu3\_3} layer of a pre-trained VGG16~\cite{simonyanVeryDeepConvolutional2015},
similar to~\cite{wuUnsupervisedLearningProbably2020}. While pixel-wise losses are common for 
autoencoders, we found it crucial to add a perceptual loss to learn textures that are 
discriminative for the pose estimation. Our full reconstruction loss can be written 
$\lrec(\img, \rec) = \lpix(\img, \rec) + \wcor\lcor(\img, \rec)$ and we use $\wcor = 10$ in 
all experiments. While our deformation-based surface parametrization naturally regularizes 
the shape, we sometimes observe bad minima where the surface has folds.  Following prior 
works~\cite{liuSoftRasterizerDifferentiable2019, chenLearningPredict3D2019, 
goelShapeViewpointKeypoints2020, zhangNeRSNeuralReflectance2021}, we thus add a small 
regularization term $\lreg = \lnorm + \llapl$ consisting of a normal consistency 
loss~\cite{desbrunImplicitFairingIrregular1999} $\lnorm$ and a Laplacian smoothing 
loss~\cite{nealenLaplacianMeshOptimization2006} $\llapl$.

\vspace{-1em}
\subsubsection{Progressive conditioning.} The goal of \textit{progressive conditioning} is to 
encourage the model to share elements (\eg, shape, texture, background) across instances to 
prevent degenerate solutions. Inspired by the curriculum learning 
philosophy~\cite{elmanLearningDevelopmentNeural1993, wangPixel2MeshGenerating3D2018, 
monnierDeepTransformationInvariantClustering2020}, we propose to do so by gradually 
increasing the latent space representing the shape, texture and background. Intuitively, 
restricting the latent space implicitly encourages maximizing the information shared across 
instances. For example, a latent space of dimension 0 (\ie, no conditioning) amounts to 
learning a global representation that is the same for all instances, while a latent space of 
dimension 1 restricts all the generated shapes, textures or backgrounds to lie on a 
1-dimensional manifold. Progressively increasing the size of the latent code during training 
can be interpreted as gradually specializing from category-level to instance-level knowledge.  
~\Cref{fig:teaser_curri} illustrates the procedure with example results where we can observe 
the progressive specialization to particular instances: reactors gradually
appear/disappear, textures get more accurate. Because common neural network implementations 
have fixed-size inputs, we implement progressive conditioning by masking, stage-by-stage, a 
decreasing number of values of the latent code. All our experiments share the same
4-stage training strategy where the latent code dimension is increased at the beginning of 
each stage and the network is then trained until convergence. We use dimensions 0/2/8/64 for 
the shape code, 2/8/64/512 for the texture code and 4/8/64/256 for the background code.  We 
provide real-image results for each stage in our supplementary material.

\vspace{-1em}
\subsubsection{Neighbor reconstruction.} The idea behind \textit{neighbor reconstruction} is 
to explicitly enforce consistency between different instances. Our key assumption is that 
neighboring instances with similar shape or texture exist in the dataset. If such neighbors 
are correctly identified, switching their shape or texture in our generation model should 
give similar reconstruction results. For a given input image, we hence propose to use its
shape or texture attribute in the image formation process of neighboring instances and apply 
our reconstruction loss on associated renderings.
Intuitively, this process can be seen as mimicking a multi-view supervision without actually 
having access to multi-view images by finding neighboring instances in well-designed latent 
spaces.  \Cref{fig:teaser_swap} illustrates the procedure with an example.

More specifically, let $\{\betadef, \genbeta, \betaaff, \bkgbeta\}$ be the
parameters predicted by our encoder for a given input training image $\img$, let $\mem$ be
a memory bank storing the images and parameters of the last $M$ instances processed by the 
network.  We write  $\mem^{\midx} = \{\img^{\midx}, \betadef^{\midx}, \genbeta^{\midx}, 
\betaaff^{\midx}, \bkgbeta^{\midx}\}$ each of these $M$ instances and associated parameters.  
We first select the closest instance from the memory bank $\mem$ in the texture (respectively 
shape) code space using the $L_2$ distance,  $m_t = \argmin_m \|\genbeta - 
\genbeta^{\midx}\|_2$ (respectively $m_s = \argmin_m \|\betadef - \betadef^{\midx}\|_2$).  We 
then swap the codes and generate the reconstruction  $\swaptx^{\midxt}$ (respectively 
$\swapsh^{\midxs}$) using the parameters $\{ \betadef^{\midxt}, \genbeta, \betaaff^{\midxt}, 
\bkgbeta^{\midxt}\}$ (respectively $\{ \betadef, \genbeta^{\midxs},\betaaff^{\midxs}, 
\bkgbeta^{\midxs}\}$).  Finally, we compute the reconstruction loss between the images 
$\img^{\midxt}$ and $\swaptx^{\midxt}$ (respectively $\img^{\midxs}$ and $\swapsh^{\midxs}$).  
Our full loss can thus be written:%
\begin{equation}
  \lswap = \lrec(\img^{\midxt}, \swaptx^{\midxt}) + \lrec(\img^{\midxs}, \swapsh^{\midxs}).
\end{equation}%
Note that we recompute the parameters of the selected instances with the current network 
state, to avoid uncontrolled effects of changes in the network state. Also note that, for 
computational reasons, we do not use this loss in the first stage where codes are almost the 
same for all instances. 

To prevent latent codes from specializing by viewpoint, we split the viewpoints into $V$ bins 
\wrt the rotation angle, uniformly sample a target bin for each input and look for the 
nearest instances only in the subset of instances within the target viewpoint range (see the 
supplementary for details).  In all experiments, we use $V = 5$ and a memory bank of size $M 
= 1024$.

\subsection{Alternate 3D and pose learning}\label{sec:optim}

Because predicting 6D poses is hard due to self-occlusions and local minima, we follow prior 
works~\cite{insafutdinovUnsupervisedLearningShape2018, hendersonLearningSingleimage3D2019, 
goelShapeViewpointKeypoints2020, tulsianiImplicitMeshReconstruction2020} and predict multiple 
pose candidates and their likelihood.  However, we identify failure modes in their 
optimization framework (detailed in our supplementary material) and instead propose a new 
optimization that alternates between 3D and pose learning.
More specifically, given an input image $\img$, we predict $K$ pose candidates $ \{(\rot_1,
\trans_1), \ldots, (\rot_K, \trans_K)\}$, and their associated probabilities 
$\probpose_{1:K}$. We render the model from the different poses, yielding $K$ reconstructions 
$\rec_{1:K}$. We then alternate the learning between two steps: (i) the \textit{3D-step} where 
shape, texture and background branches of the network are updated by minimizing $\lmax$ using 
the pose associated to the highest probability,
and (ii) the \textit{P-step} where the branches of the network predicting candidate
poses and their associated probabilities are updated by 
minimizing:%
\begin{equation}\label{eq:lexp}
  \textstyle
  \lexp = \sum_{k}\probpose_k \lrec(\img, \rec_k) + \wpose\lpose,
\end{equation}%
where $\lrec$ is the reconstruction loss described 
in~\cref{sec:learning}, $\lpose$ is a regularization loss on the predicted poses and $\wpose$ 
is a scalar hyperparameter. More precisely, we use $\lpose = \sum_k |\bar{\probpose}_k - 
\nicefrac{1}{K}|$ where
$\bar{\probpose}_k$ is the averaged probabilities for candidate $k$ in a particular training 
batch. Similar to~\cite{hendersonLearningSingleimage3D2019}, we found it crucial to introduce 
this regularization term to encourage the use of all pose candidates. In particular, this 
prevents a collapse mode where only few pose candidates are used. Note that we do not use the 
neighbor reconstruction loss nor the mesh regularization loss which are not relevant for 
viewpoints.  In all experiments, we use $\wpose = 0.02$.
  
Inspired by the camera multiplex of~\cite{goelShapeViewpointKeypoints2020}, we parametrize 
rotations with the classical Euler angles (azimuth, elevation and roll) and rotation 
candidates correspond to offset angles \wrt reference viewpoints.
Since in practice elevation has limited variations, our reference viewpoints are uniformly 
sampled along the azimuth dimension. Note that compared 
to~\cite{goelShapeViewpointKeypoints2020}, we do not directly optimize a set of pose 
candidates per training image, but instead learn a set of $K$ predictors for the entire 
dataset. We use $K = 6$ in all experiments.

\section{Experiments}\label{sec:exp}
\vspace{-.5em}

We validate our approach in two standard setups. It is first quantitatively evaluated on 
ShapeNet where state-of-the-art methods use multiple views as supervision.  Then, we compare 
it on standard real-image benchmarks and demonstrate its applicability to more complex 
datasets. Finally, we present an ablation study.

\subsection{Evaluation on the ShapeNet benchmark}

We compare our approach to state-of-the-art methods using multi-views, viewpoints and 
silhouettes as supervision. Our method is instead learned without supervision. For all 
compared methods, one model is trained per class. We adhere to community 
standards~\cite{katoNeural3DMesh2018, liuSoftRasterizerDifferentiable2019, 
niemeyerDifferentiableVolumetricRendering2020} and use the renderings and splits 
from~\cite{katoNeural3DMesh2018} of the ShapeNet 
dataset~\cite{changShapeNetInformationRich3D2015}. It corresponds to a subset of 13 classes 
of 3D objects, each object being rendered into a $64\times64$ image from 24 viewpoints 
uniformly spaced along the azimuth dimension. We evaluate all methods using the standard 
Chamfer-$L_1$ distance~\cite{meschederOccupancyNetworksLearning2019, 
niemeyerDifferentiableVolumetricRendering2020}, where predicted shapes are pre-aligned using 
our gradient-based version of the Iterative Closest Point 
(ICP)~\cite{beslMethodRegistration3D1992} with anisotropic scaling. Indeed, compared to 
competing methods having access to the ground-truth viewpoint during training, we need to 
predict it for each input image in addition to the 3D shape. This yields to both shape/pose 
ambiguities (\eg, a small nearby object or a bigger one far from the camera) and small 
misalignment errors that dramatically degrade the performances. We provide evaluation details 
as well as results without ICP in our supplementary.

\begin{figure}[t]
  \begin{floatrow}
    \capbtabbox[0.6\columnwidth]{
  \centering
  \scriptsize
  \addtolength{\tabcolsep}{2pt}
  \begin{tabular}{@{}l|ccc|c@{}}
  \toprule
  Method & \bf Ours & SDF-SRN~\cite{linSDFSRNLearningSigned2020} & 
  DVR~\cite{niemeyerDifferentiableVolumetricRendering2020} & 
  DVR~\cite{niemeyerDifferentiableVolumetricRendering2020} \\
  \mv & & & & \cmark \\
  \cam & &\cmark & \cmark & \cmark\\
  \sil &  & \cmark & \cmark & \cmark\\

  \midrule

  airplane &\bf 0.110 & 0.128 & 0.114 & \bf 0.111 \\
  bench &\bf 0.159  & - & 0.255 & \bf 0.176 \\
  cabinet &\bf 0.137  & - & 0.254 & \bf 0.158 \\
  car & 0.168  & \bf 0.150 & 0.203 & \bf 0.153 \\
  chair &\bf 0.253 &  0.262 & 0.371 & \bf 0.205 \\
  display &\bf 0.220 & - & 0.257 & \bf 0.163 \\
  lamp &\red{0.523} & - & \bf 0.363 & \bf 0.281 \\
  phone &\bf 0.127 & - & 0.191 & \bf 0.076 \\
  rifle &\bf 0.097 & - & 0.130 & \bf 0.083 \\
  sofa &\bf 0.192 & - & 0.321 & \bf 0.160 \\
  speaker &\bf 0.224 & - & 0.312 & \bf 0.215 \\
  table &\bf 0.243 & - & 0.303 & \bf 0.230 \\
  vessel &\bf 0.155 & - & 0.180 & \bf 0.151 \\

  \midrule
  
  mean  &\bf 0.201 & - & 0.250 & \bf 0.166 \\

  \bottomrule
  \end{tabular}

}{\vspace{-.5em}
      \caption{\textbf{ShapeNet comparison.} We report Chamfer-$L_1\downarrow$ obtained after 
      ICP, \textbf{best} results are highlighted. Supervisions are: \multiview, 
    \camerakeypoint, \silhouette.}
      \label{tab:shapenet}}\,
  \ffigbox[0.36\columnwidth]{\centering\includegraphics[width=\columnwidth]{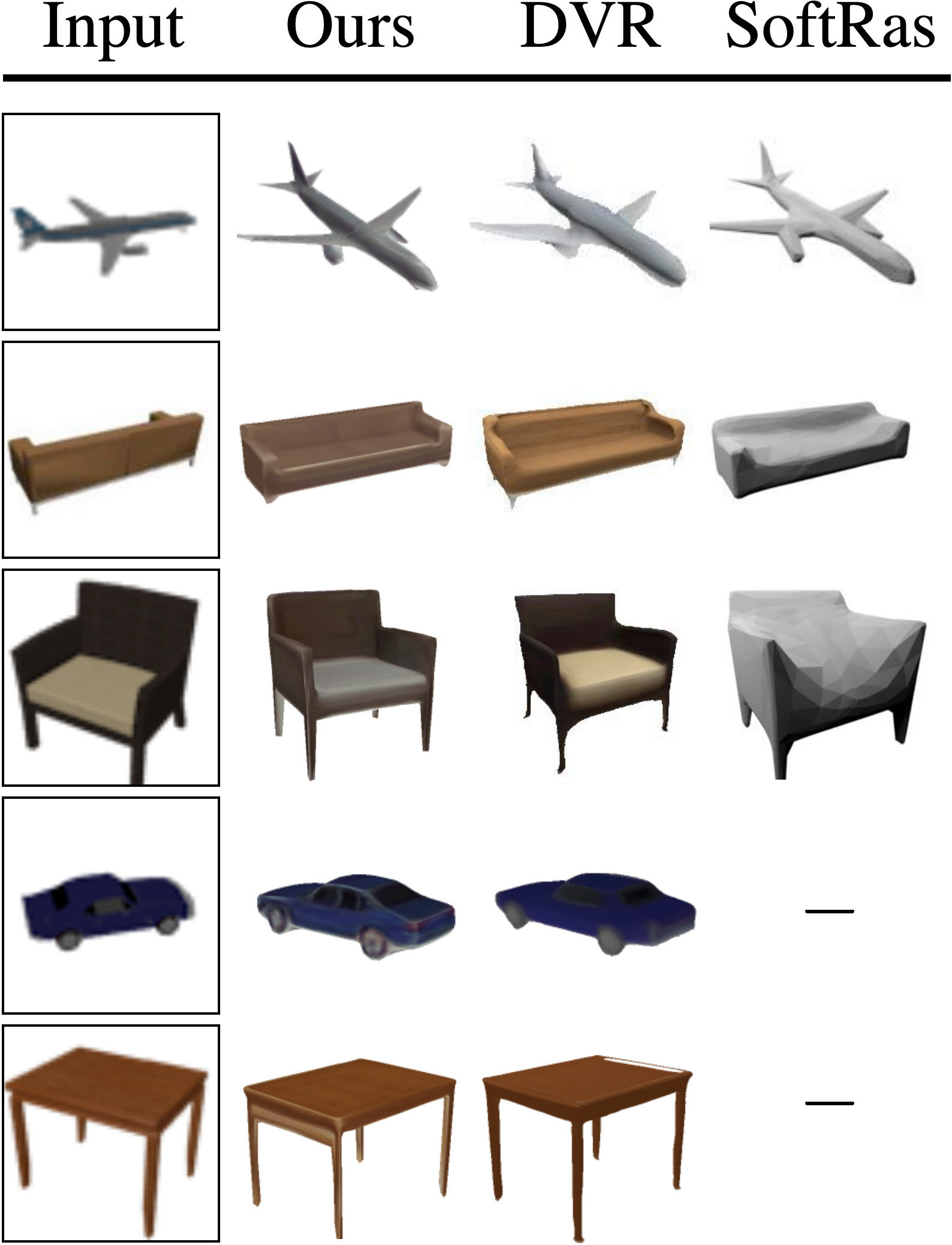}}{%
    \vspace{-.5em}
    \caption{\textbf{Visual comparisons.} We compare to
      DVR~\cite{niemeyerDifferentiableVolumetricRendering2020} and
      SoftRas~\cite{liuSoftRasterizerDifferentiable2019} learned with full supervision (\mv, 
      \cam, \sil).
  }\label{fig:shapenet}}
  \vspace{-1.2em}
  \end{floatrow}
\end{figure}

We report quantitative results and compare to the state of the art
in~\Cref{tab:shapenet}, where methods using multi-views are visually separated. We evaluate 
the pre-trained weights for SDF-SRN~\cite{linSDFSRNLearningSigned2020} and train the models 
from scratch using the official implementation for 
DVR~\cite{niemeyerDifferentiableVolumetricRendering2020}. We tried evaluating
SMR~\cite{huSelfSupervised3DMesh2021} but could not reproduce the results. We do not 
compare to TARS~\cite{duggalTopologicallyAwareDeformationFields2022} which is based on 
SDF-SRN and share the same performances. Our approach achieves results that are on average 
better than the state-of-the-art methods supervised with silhouette and viewpoint 
annotations. This is a strong result: while silhouettes are trivial in this benchmark, 
learning without viewpoint annotations is extremely challenging as it involves solving
the pose estimation and shape reconstruction problems simultaneously. For some categories, 
our performances are even better than DVR supervised with multiple views. This shows that our 
system learned on raw images generates 3D reconstructions comparable to the ones obtained 
from methods using geometry cues like silhouettes and multiple views.  Note that for the lamp 
category, our method predicts degenerate 3D shapes; we hypothesize this is due to their
rotation invariance which makes the viewpoint estimation ambiguous.

We visualize and compare the quality of our 3D reconstructions in~\Cref{fig:shapenet}.
The first three examples correspond to examples advertised in 
DVR~\cite{niemeyerDifferentiableVolumetricRendering2020}, the last two corresponds to examples 
we selected. Our method generates textured meshes of high-quality across all these 
categories. The geometry obtained is sharp and accurate, and the predicted texture mostly 
corresponds to the input.

\subsection{Results on real images}

\subsubsection{Pascal3D+ Car and CUB benchmarks.}
We compare our approach to state-of-the-art SVR methods on real images, where multiple views 
are not available. All competing methods use silhouette supervision and output meshes that 
are symmetric. CMR~\cite{kanazawaLearningCategorySpecificMesh2018} additionally use 
keypoints, UCMR~\cite{goelShapeViewpointKeypoints2020} and 
IMR~\cite{tulsianiImplicitMeshReconstruction2020} starts learning from a given template 
shape; we \textit{do not} use any of these and directly learn from raw images. We strictly 
follow the community standards~\cite{kanazawaLearningCategorySpecificMesh2018, 
goelShapeViewpointKeypoints2020, tulsianiImplicitMeshReconstruction2020} and use the 
train/test splits of Pascal3D+ Car~\cite{xiangPASCALBenchmark3D2014} (5000/220 images) and 
CUB-200-2011~\cite{welinderCaltechucsdBirds2002010} (5964/2874 images). Images are 
square-cropped around the object using bounding box annotations and resized to $64\times64$.

\begin{table}[t]
  \centering
  \scriptsize
  \addtolength{\tabcolsep}{2pt}
  \renewcommand{\arraystretch}{.95}
  \begin{tabular}{@{}lcccccccc}
  \toprule
  & \multicolumn{3}{c}{Supervision} & \multicolumn{3}{c}{Pascal3D+ Car} & 
  \multicolumn{2}{c}{CUB-200-2011}\\

  \cmidrule(lr){2-4}\cmidrule(lr){5-7} \cmidrule(lr){8-9}
 
  Method & \ck & \sil & \pri & 3D IoU $\uparrow$ & Ch-$L_1$ $\downarrow$ & Mask IoU 
  $\uparrow$ & PCK@0.1 $\uparrow$ & Mask IoU $\uparrow$\\

  \midrule
  CMR~\cite{kanazawaLearningCategorySpecificMesh2018} & \cmark & \cmark & \cmark & 64 & - & - 
  & 48.3 & 70.6\\
  IMR~\cite{tulsianiImplicitMeshReconstruction2020} & & \cmark & \cmark & - & - & - & 53.5 & 
  -\\
  UMR~\cite{liSelfsupervisedSingleview3D2020} &  & \cmark & \cmark & 62 & - & - & 58.2 & 
  73.4\\
  UCMR~\cite{goelShapeViewpointKeypoints2020} & & \cmark & \cmark &\bf 67.3 & 0.172 & 73.7 & 
  - & 63.7\\
  SMR~\cite{huSelfSupervised3DMesh2021} &  & \cmark & \cmark & - & - & - &\bf 62.2 & \bf 
  80.6\\

  \textbf{Ours} & & & & 65.9 &\bf 0.163 &\bf 83.9 & 49.0 & 71.4\\
  \bottomrule
  \end{tabular}

  \vspace{-.8em}
  \caption{\textbf{Real-image quantitative comparisons.} Supervision corresponds to 
  \camerakeypoint, \silhouette, \prior (see~\cref{tab:comp} for details).  \vspace{-.5em}}
  \label{tab:real}
\end{table}

\begin{figure}[t]
  \centering
  \includegraphics[width=\columnwidth]{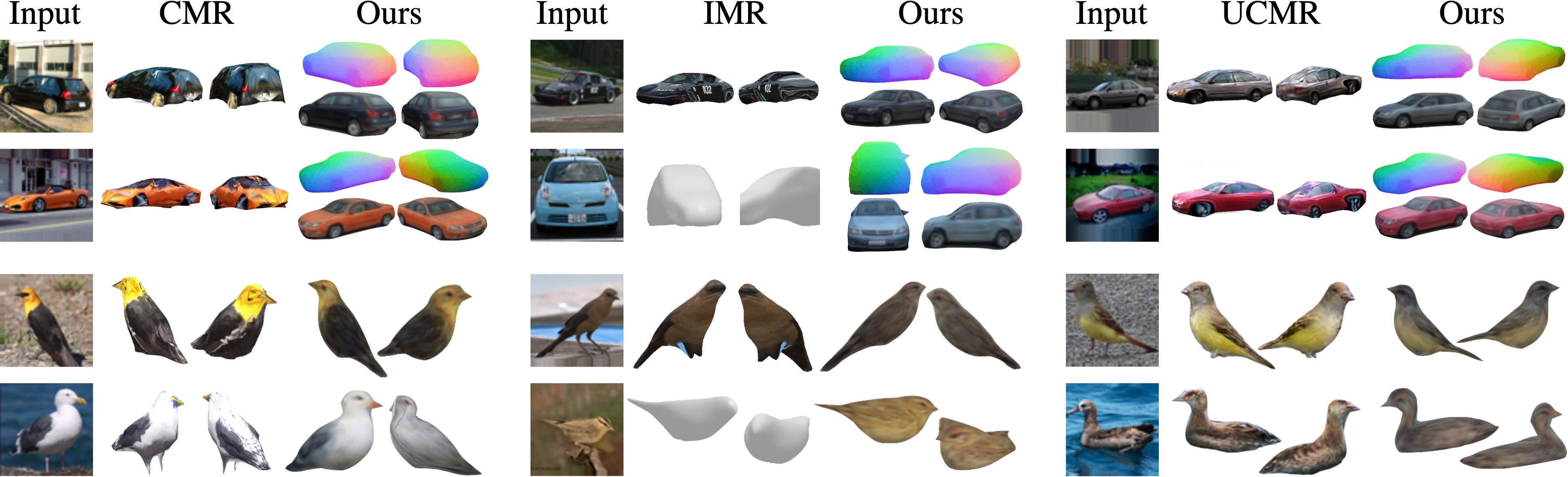}
  \caption{\textbf{Real-image comparisons.} We show reconstructions on Pascal3D+ Cars (top) 
    and CUB (bottom) and compare to CMR~\cite{kanazawaLearningCategorySpecificMesh2018},
  IMR~\cite{tulsianiImplicitMeshReconstruction2020},
  UCMR~\cite{goelShapeViewpointKeypoints2020}. \vspace{-.8em}}
  \vspace{-.5em}
  \label{fig:real_comp}
\end{figure}

A quantitative comparison is summarized in~\Cref{tab:real}, where we report 3D IoU, 
Chamfer-$L_1$ (with ICP alignment), Mask IoU for Pascal3D+ Car, and Percentage of Correct 
Keypoints thresholded at $\alpha = 0.1$ (PCK@0.1)~\cite{kulkarniCanonicalSurfaceMapping2019}, 
Mask IoU for CUB. Our approach yields competitive results across all metrics although it does 
not rely on any supervision used by other works.  On Pascal3D+ Car, we achieve significantly 
better results than UCMR for Chamfer-$L_1$ and Mask IoU, which we argue are less biased 
metrics than the standard 3D IoU~\cite{tulsianiMultiviewSupervisionSingleview2017, 
kanazawaLearningCategorySpecificMesh2018} computed on unaligned shapes (see supplementary).  
On CUB, our approach achieves reasonable results that are however slightly worse than the 
state of the art. We hypothesize this is linked to our pose regularization term encouraging 
the use of all viewpoints whereas these bird images clearly lack back views.

We qualitatively compare our approach to the state of the art in~\Cref{fig:real_comp}. For 
each input, we show the mesh rendered from two viewpoints. For our car results, we 
additionally show meshes with synthetic textures to emphasize correspondences.  
Qualitatively, our approach yields results on par with prior works both in terms of geometric 
accuracy and overall realism. Although the textures obtained 
in~\cite{tulsianiImplicitMeshReconstruction2020} look more accurate, they are modeled as
pixel flows, which has a clear limitation when synthesizing unseen texture parts. Note that 
we do not recover details like the bird legs which are missed by prior works due to coarse 
silhouette annotations. We hypothesize we also miss them because they are hardly consistent 
across instances, \eg, legs can be bent in multiple ways.

\begin{figure}[t]
    \centering
    \begin{subfigure}[t]{0.501\textwidth}
      \centering
      \includegraphics[width=\columnwidth]{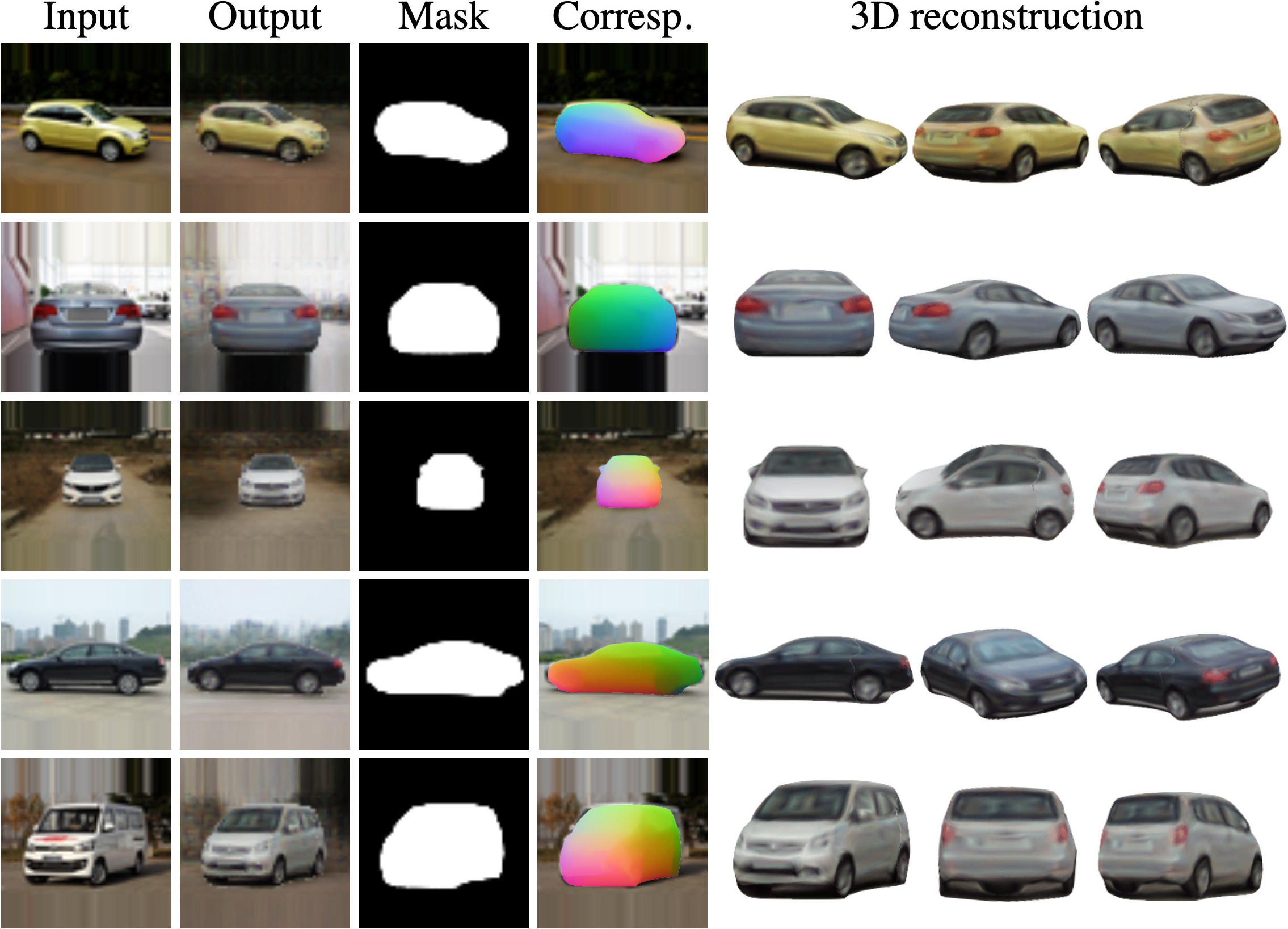}
      \vspace{-1em}
      \caption{CompCars~\cite{yangLargescaleCarDataset2015}}
      \label{fig:compcar}
    \end{subfigure}\hspace*{\fill}
    \begin{subfigure}[t]{0.462\textwidth}
      \centering
      \includegraphics[width=\columnwidth]{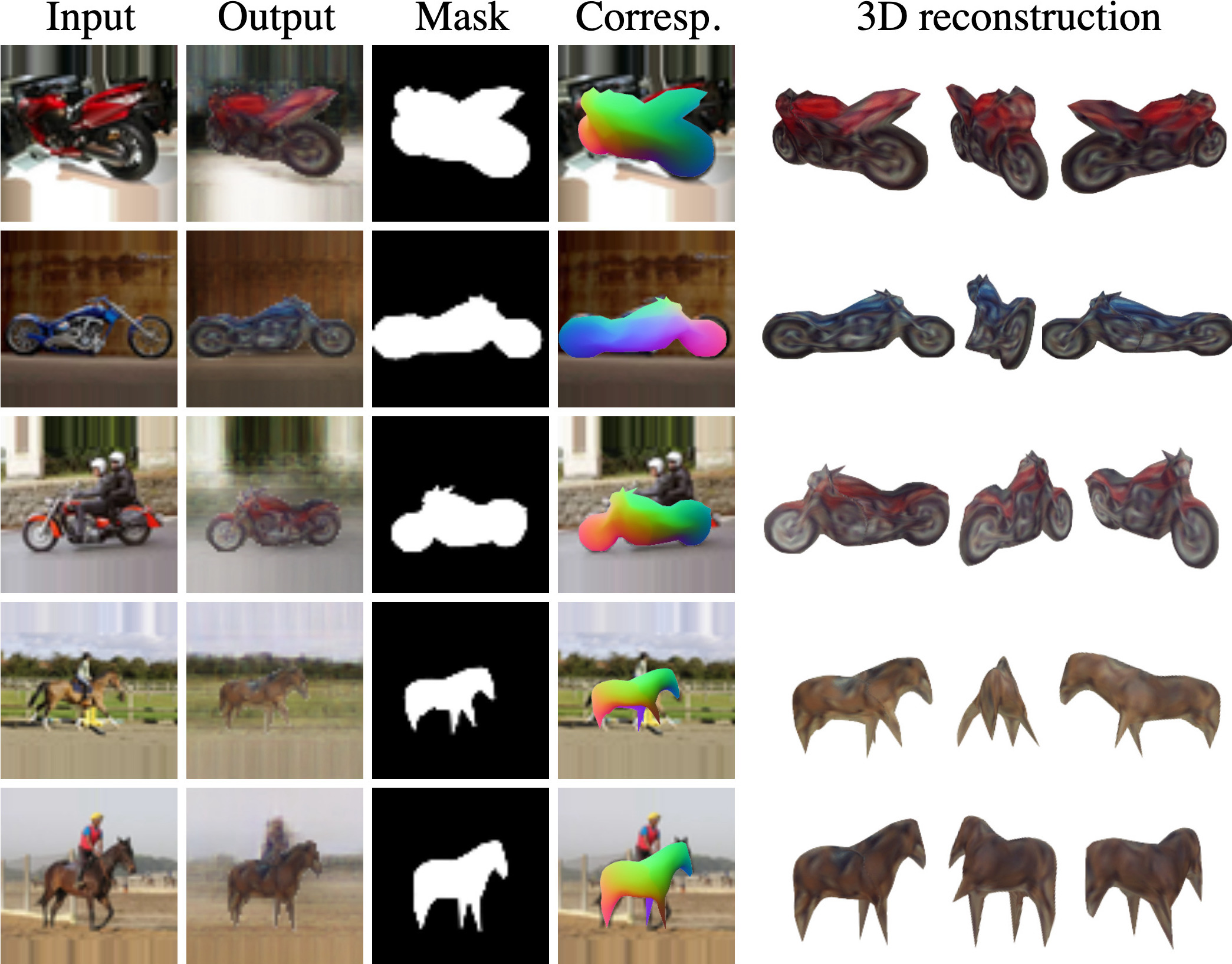}
      \vspace{-1em}
      \caption{LSUN Motorbike and Horse~\cite{yuLSUNConstructionLargescale2016}}
      \label{fig:lsun}
    \end{subfigure}
    \caption{\textbf{Real-world dataset results.} From left to right, we show: input and 
      output images, the predicted mask, a correspondence map and the mesh rendered from 3
      viewpoints. Note that for LSUN Horse, the geometry quality is low and outlines our
approach limitations (see text). Best viewed digitally. \vspace{-.2em}}
    \label{fig:realworld}
\end{figure}

\vspace{-1em}
\subsubsection{Real-word datasets.} Motivated by 3D-aware image synthesis methods learned
in-the-wild~\cite{nguyen-phuocHoloGANUnsupervisedLearning2019, 
niemeyerGIRAFFERepresentingScenes2021}, we investigate whether our approach can be applied to 
real-world datasets where silhouettes are not available and images are not methodically 
cropped around the object.  We adhere to standards from the 3D-aware image synthesis 
community~\cite{nguyen-phuocHoloGANUnsupervisedLearning2019, 
niemeyerGIRAFFERepresentingScenes2021} and apply our approach to $64\times64$ images of 
CompCars~\cite{yangLargescaleCarDataset2015}. In addition, we provide results for the more 
difficult scenario of LSUN images~\cite{yuLSUNConstructionLargescale2016} for motorbikes and 
horses. Because many LSUN images are noise, we filter the datasets as follows: we manually 
select 16 reference images with different poses, find the nearest neighbors from the first 
200k images in a pre-trained ResNet-18~\cite{heDeepResidualLearning2016} feature space, and 
keep the top 2k for each reference image. We repeat the procedure with flipped reference 
images yielding 25k images.

Our results are shown in~\Cref{fig:realworld}. For each input image, we show from left to 
right: the output image, the predicted mask, a correspondence map, and the 3D reconstruction 
rendered from the predicted viewpoint and two other viewpoints. Although our approach is 
trained to synthesize images, these are all natural by-products.
While the quality of our 3D car reconstructions is high, the reconstructions obtained for 
LSUN images lack some realism and accuracy (especially for horses), thus outlining
limitations of our approach. However, our segmentation and correspondence maps emphasize our 
system ability to accurately localize the object and find correspondences, even when the 
geometry is coarse.

\vspace{-1em}
\subsubsection{Limitations.} Even if our approach is a strong step towards generic 
unsupervised SVR, we can outline three main limitations. First, the lack of different views in 
the data harms the results (\eg, most CUB birds have concave backs); this can be linked to 
our uniform pose regularization term which is not adequate in these cases. Second, complex 
textures are not predicted correctly (\eg, motorbikes in LSUN). Although we argue it could be 
improved by more advanced autoencoders, the neighbor reconstruction term may prevent unique 
textures to be generated.  Finally, despite its applicability to multiple object categories 
and diverse datasets, our multi-stage progressive training is cumbersome and an automatic way 
of progressively specializing to instances is much more desirable.

\begin{figure}[t]
  \begin{floatrow}
    \capbtabbox[0.356\columnwidth]{
  \centering
  \scriptsize
  \addtolength{\tabcolsep}{2pt}
  \begin{tabular}{@{}lccc@{}}
  \toprule
  Model & Full & w/o & w/o\\ 
  & & $\lswap$ & PC\\ 
  \midrule
  airplane & 0.110 & 0.124 &\bf 0.107\\
  bench &\bf 0.159 & 0.188 & 0.206\\
  car & \bf 0.168 & 0.179 & 0.173\\
  chair &\bf 0.253 & 0.319 & 0.527\\
  table &\bf 0.243 & 0.246 & 0.598\\
  \midrule
  mean & \bf 0.187 & 0.211 & 0.322\\
  \bottomrule
  \end{tabular}
}{
  \caption{\textbf{Ablation results on ShapeNet{\normalfont 
\cite{changShapeNetInformationRich3D2015}}.}}\label{tab:ablation}}\;
  \ffigbox[0.56\columnwidth]{\centering\includegraphics[width=\columnwidth]{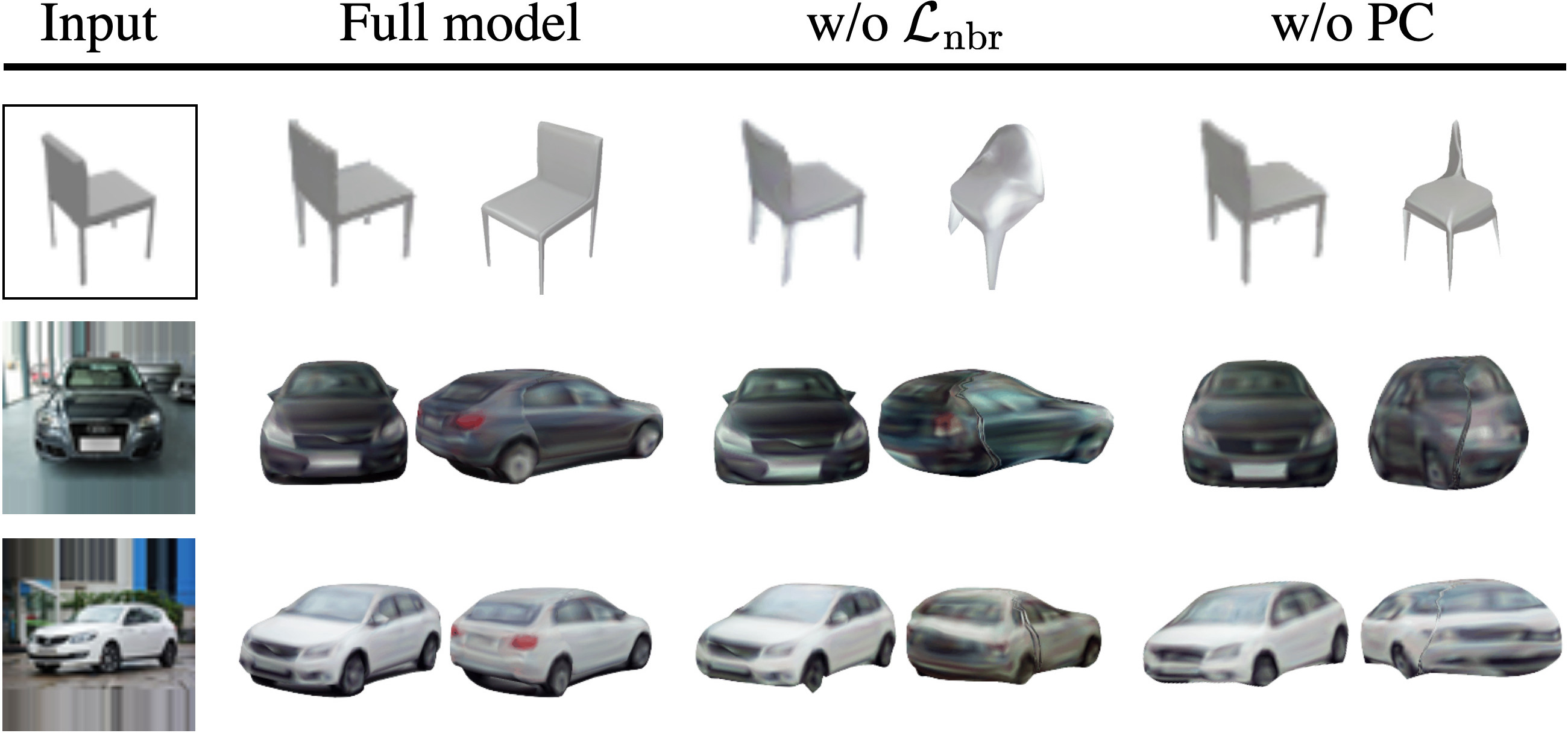}
  }{%
    \caption{\textbf{Ablation visual results.} For each input, we show the mesh rendered from 
  two viewpoints.}\label{fig:ablation}}
  \end{floatrow}
\end{figure}

\subsection{Ablation study}

We analyze the influence of our neighbor reconstruction loss $\lswap$ and progressive 
conditioning (PC) by running experiments without each component.

First, we provide quantitative results on ShapeNet in~\Cref{tab:ablation}. When $\lswap$ is 
removed, the results are worse for almost all categories, outlining that it is important to 
the predicted geometry accuracy. When PC is removed, results are comparable to the full model 
for airplane and car but much worse for chair and table. Indeed, they involve more complex 
shapes and our system falls into a bad minimum with degenerate solutions, a scenario that is 
avoided with PC.

Second, we perform a visual comparison on ShapeNet and CompCars examples 
in~\Cref{fig:ablation}.  For each input, we show the mesh rendered from the predicted 
viewpoint and a different viewpoint. When $\lswap$ is removed, we observe that the 
reconstruction seen from the predicted viewpoint is correct but it is either wrong for chairs 
and degraded for cars when seen from the other viewpoint. Indeed, the neighbor reconstruction 
explicitly enforces the unseen reconstructed parts to be consistent with other instances.  
When PC is removed, we observe degenerate reconstructions where the object seen from a 
different viewpoint is not realistic.

\section{Conclusion}

We presented UNICORN, an unsupervised SVR method which, in contrast to all prior works, 
learns from raw images only. We demonstrated it yields
high-quality results for diverse shapes as well as challenging real-world image collections.  
This was enabled by two key contributions aiming at leveraging consistency across different 
instances: our {\it progressive conditioning} training strategy and {\it neighbor
reconstruction} loss. We believe our work includes both an important step forward for 
unsupervised SVR and the introduction of a valuable conceptual insight.

\vspace{-.8em}
\subsubsection{Acknowledgements.} We thank Fran\c{c}ois Darmon for inspiring discussions; 
Robin Champenois, Romain Loiseau, Elliot Vincent for manuscript feedback; Michael Niemeyer, 
Shubham Goel for evaluation details. This work was supported in part by ANR project EnHerit 
ANR-17-CE23-0008, project Rapid Tabasco, gifts from Adobe and HPC resources from GENCI-IDRIS 
(2021-AD011011697R1).




\bibliographystyle{splncs04}
\bibliography{references}

\clearpage
\appendix

{\Large\bf\centering Supplementary Material for\\
  Share With Thy Neighbors: Single-View\\ Reconstruction by Cross-Instance 
Consistency\\[1.5em]}

In this supplementary document, we first describe our custom differentiable rendering 
function (\Cref{sec:diff}). Then, we present additional model insights (\Cref{sec:model}) as 
well as quantitative evaluation details (\Cref{sec:eval}). Finally, we provide implementation 
details~(\Cref{sec:implem}), including network architectures, design choices and training 
details. 

\section{Custom differentiable rendering}\label{sec:diff}

Our output images correspond to the soft rasterization of a textured mesh on top of a 
background image.  We observe divergence results when learning geometry from raw photometric 
comparison with the standard SoftRasterizer~\cite{liuSoftRasterizerDifferentiable2019} and 
thus propose two key  changes. In the following, given a mesh $\mesh$ and a background 
$\bkg$, we describe our rendering function $\rend$ producing the image $\rec = \rend(\mesh, 
\bkg)$.  We first present SoftRasterizer formulation and its limitations, then introduce our 
modifications. In the following, we write pixel-wise multiplication with $\odot$ and the 
division of image-sized tensors corresponds to pixel-wise division.

\vspace{-1em}
\subsubsection{SoftRasterizer formulation.} Given a 2D pixel location $\pixi$, the influence 
of a face $\facej$ is modeled by an occupancy function:%
\begin{equation}
  \occsr(\pixi, \facej) = \sigmoid\Big(\frac{\dist(\pixi, \facej)}{\sigma}\Big),
\end{equation}%
%
where $\sigma$ is a temperature, $\dist(\pixi, \facej)$ is the signed Euclidean distance 
between pixel $\pixi$ and projected face $\facej$. Let us call $L -1$ the maximum number of 
faces intersecting the ray associated to a pixel and sort, for each pixel, the intersecting 
faces by increasing depth. Image-sized maps for occupancy $\prob_\ell$, color $\col_\ell$ and 
depth $\depth_\ell$ are built associating to each pixel the $\ell$-th intersecting face 
attributes. Background is modeled as additional maps such that $\prob_L = 1, \col_L = \bkg$, 
and $\depth_L = \zbkg$ is a constant, far from the camera. The SoftRasterizer's aggregation 
function $\aggsr$ merges them to render the final image $\rec$:
\begin{equation}\label{eq:agg_sr}
  \aggsr(\prob_{1:L}, \col_{1:L}, \depth_{1:L}) = \sum_{\ell = 1}^L \frac{\prob_\ell \odot
  \exp(\depthinv_\ell / \gamma)}{\sum_k \prob_k \odot \exp(\depthinv_k / \gamma)}\; \odot 
  \col_\ell,
\end{equation}%
where $\gamma$ is a temperature parameter, $\depthinv_\ell = \frac{\zfar - \depth_\ell}{\zfar 
- \znear}$ and $\znear, \zfar$ correspond to near/far cut-off distances.  This formulation 
hence relies on 5 hyperparameters ($\sigma$, $\gamma$, $\znear$, $\zfar$, $\zbkg$) and 
default values are $\sigma = \gamma = 10^{-4}$, $\znear = 1$, $\zfar = 100$ and $\frac{\zfar 
- \zbkg}{\zfar - \znear} = \epsilon = 10^{-3}$.

The formulation introduced in~\Cref{eq:agg_sr} has one main limitation: gradients don't flow 
well through $\prob_{1:L}$ obtained by soft rasterization, and thus vertex positions cannot 
be optimized by raw photometric comparison. The simple case of a single face on a black 
background gives:%
\begin{equation}
  \rec = \frac{\prob_1 \odot e^{\nicefrac{\depthinv_1}{\gamma}}}{\prob_1 \odot 
  e^{\nicefrac{\depthinv_1}{\gamma}} + e^{\nicefrac{\epsilon}{\gamma}}}\; \odot \col_1
  \approx \frac{\prob_1 \odot e^{\nicefrac{\depthinv_1}{\gamma}}}{\prob_1 \odot
  e^{\nicefrac{\depthinv_1}{\gamma}}}\; \odot \col_1 = \col_1,
\end{equation}%
for almost all $\prob_1, \depthinv_1$. Indeed, considering $x, \eta > 0$, we have $x 
e^{\nicefrac{(\epsilon + \eta)}{\gamma}} \gg e^{\nicefrac{\epsilon}{\gamma}}$ iif $x \gg 
e^{\nicefrac{-\eta}{\gamma}}$. Even in the best case where $\eta = \epsilon = 10^{-3}$ (\ie, 
the object is close to $\zfar$), this holds for all $x \gg e^{-10} \approx 4\times10^{-5}$!  
We found that tuning $\gamma$ was not sufficient to mitigate the issue, one would have to 
tune $\gamma, \znear, \zfar, \zbkg$ simultaneously to enable the optimization of the vertex 
positions.

\vspace{-1em}
\subsubsection{Our layered formulation.} Inspired by layered image 
models~\cite{jojicLearningFlexibleSprites2001, monnierUnsupervisedLayeredImage2021}, we 
propose to model the rendering of a mesh as the layered composition of its projected face 
attributes. More specifically, given occupancy $\prob_{1:L}$ and color $\col_{1:L}$ maps, we 
render an image $\rec$ through the classical recursive alpha compositing:%
\begin{equation}
  \agg(\prob_{1:L}, \col_{1:L}) = \sum_{\ell = 1}^L \Big[\prod_{k<\ell}^L(1 - \prob_k)\Big] 
  \odot \prob_\ell \odot \col_\ell.
\end{equation}%
This formulation has a clear interpretation where color maps are overlaid on top of each 
other with a transparency corresponding to their occupancy map. Note that we choose to drop 
the explicit depth dependency as all 3D coordinates (including depth) of a vertex already 
receive gradients by 3D-to-2D projection. Our layered aggregation used together with the 
SoftRasterizer's occupancy function $\occsr$ results in face inner-borders that are visually 
unpleasant. We thus instead use the occupancy function introduced 
in~\cite{chenLearningPredict3D2019} defined by:%
\begin{equation}
  \occ(\pixi, \facej) = \exp(\min(0, \frac{\dist(\pixi, \facej)}{\sigma})).
\end{equation}%
Compared to $\occsr$, this function yields constant occupancy of 1 inside the faces.  In 
addition to its simplicity, our differential renderer has two main advantages compared to 
SoftRasterizer.  First, gradients can directly flow through occupancies $\prob_{1:L}$ and the 
vertex positions can be updated by photometric comparison. Second, our formulation involves 
only one hyperparameter ($\sigma$) instead of five, making it easier to use.

\section{Model insights}\label{sec:model}

\subsection{Progressive conditioning}

\begin{figure}[t]
    \centering
    \includegraphics[width=\columnwidth]{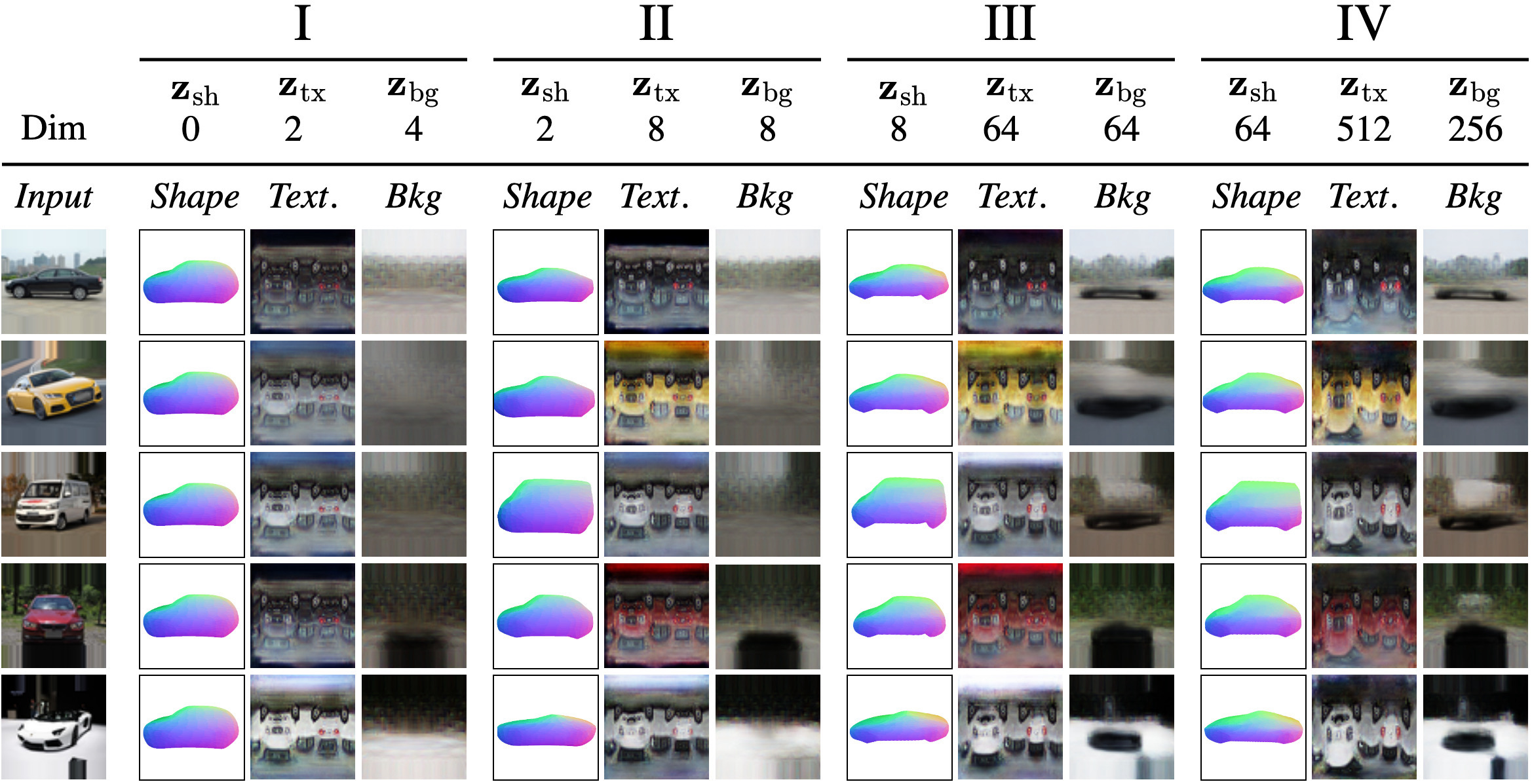}
    \caption{\textbf{Progressive conditioning on 
    CompCars~{\normalfont\cite{yangLargescaleCarDataset2015}}.} Given an input image 
    (leftmost column), we show for each training stage, from left to right, a side view of 
  the predicted shape, the texture image and the background image.\vspace{-1em}}
    \label{fig:curri_full}
\end{figure}

\Cref{fig:curri_full} shows the results obtained on 
CompCars~\cite{yangLargescaleCarDataset2015} at the end of each stage of the training. Given 
an input image (leftmost column), we show for each training stage the predicted outputs.  
From left to right, they correspond to a side view of the shape, the texture image and the 
background image.  We can observe that all shape, texture and background models gradually 
specialize to the instance represented in the input.  In particular, this allows us to start 
with a weak background model to avoid degenerate solutions and to end up with a powerful 
background model to improve the reconstruction quality. Also note how all the texture images 
are aligned.

\subsection{Neighbor reconstruction}

When computing the neighbor reconstructions, we explicitly find neighbors that have a 
viewpoint different from the predicted viewpoint. More specifically, for a given input, we 
compute the angle between the predicted rotation matrix and all rotation matrices of the 
memory bank.  Following standard conventions, such an angle lies in $[0^\circ, 180^\circ]$.  
Then, we select a target angle range as follows: we split the range of angles $[20^\circ, 
180^\circ]$ into a partition of $V$ uniform and continuous bins, and we uniformly sample one 
of the $V$ angle ranges.  Finally, we look for neighbors in the subset of instances having an 
angle within the selected range. In all experiments, we use $V = 5$.

We use a total angle range of $[20^\circ, 180^\circ]$ instead of $[0^\circ, 180^\circ]$ to 
remove instances having a similar pose. Note that we first tried to find neighbors of
different poses without further constraint (which amounts to using $V = 1$) but we found that 
learned latent codes were specialized by viewpoints, \eg, front / back view images 
corresponding to a shape mode with unrealistic side views, and side view images corresponding 
to a shape mode with unrealistic front / back views.

\subsection{Joint 3D and pose learning}

\begin{figure}[t]
    \centering
    \begin{subfigure}[t]{0.384\textwidth}
      \includegraphics[width=\columnwidth]{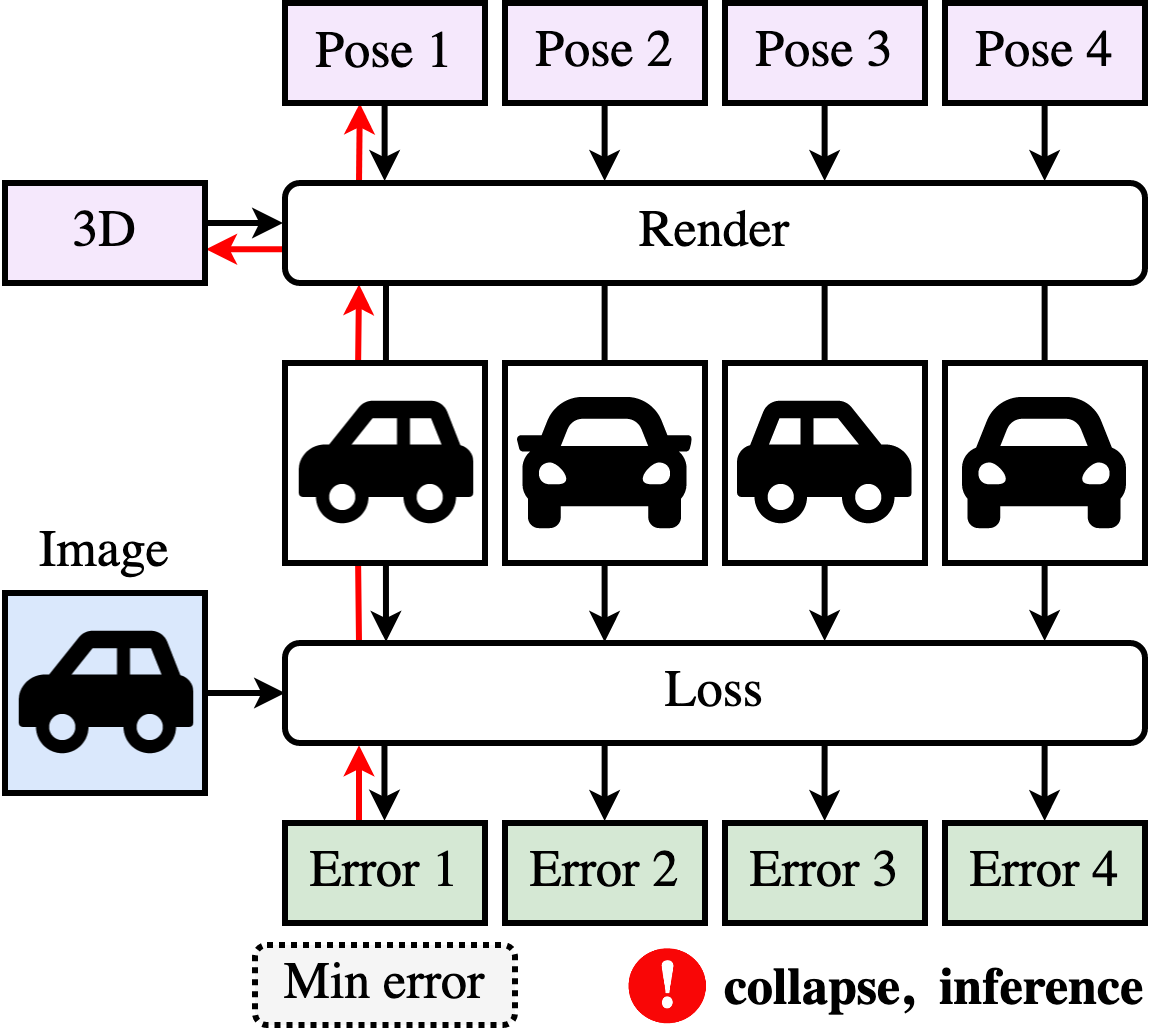}
      \caption{Optimization in~\cite{insafutdinovUnsupervisedLearningShape2018}}
      \label{fig:opt_prior1}
    \end{subfigure}\hspace*{\fill}
    \begin{subfigure}[t]{0.563\textwidth}
      \centering
      \includegraphics[width=\columnwidth]{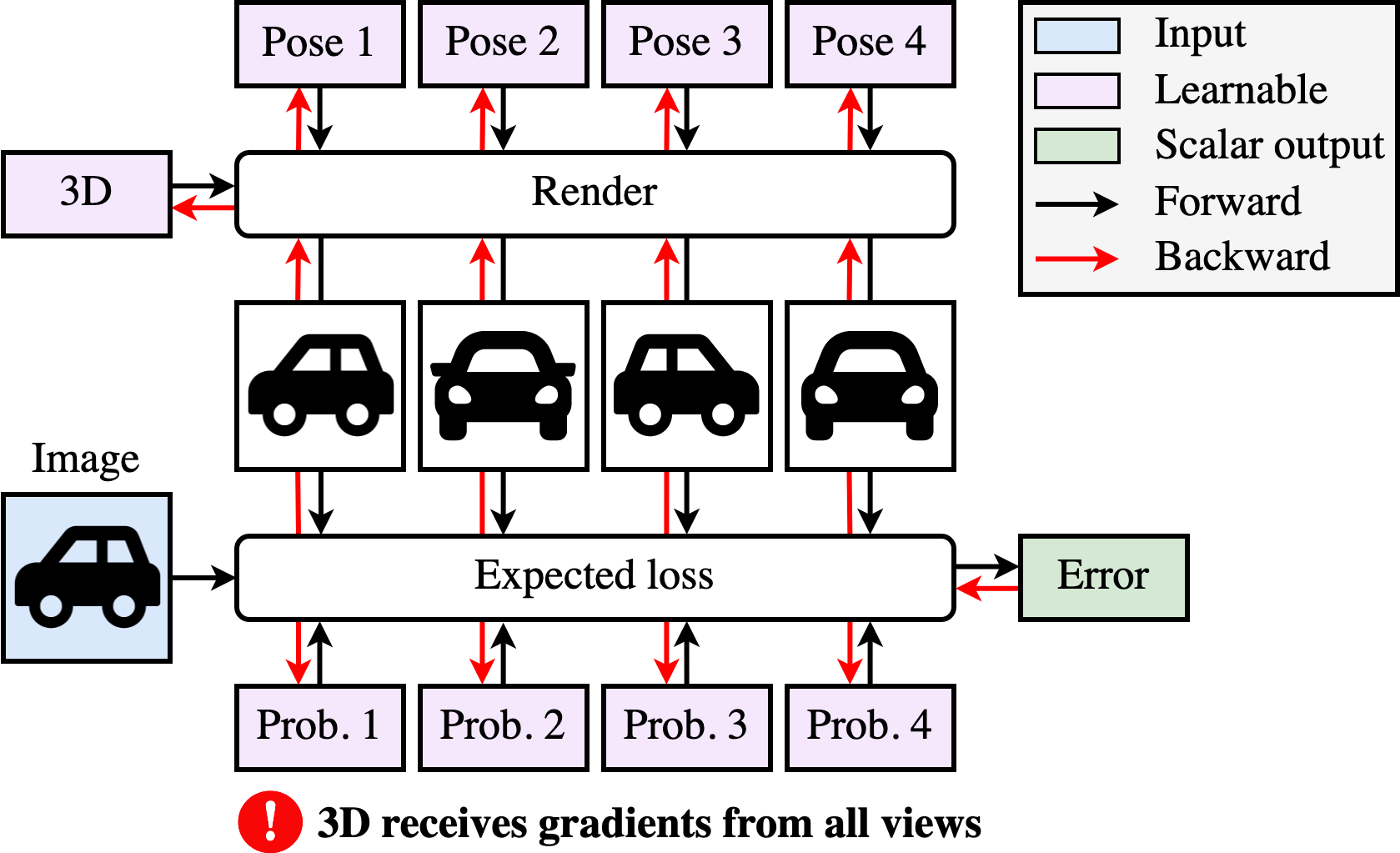}
      \caption{Optimization in~\cite{hendersonLearningSingleimage3D2019, 
      goelShapeViewpointKeypoints2020, tulsianiImplicitMeshReconstruction2020}}
    \label{fig:opt_prior2}
    \end{subfigure}
    \caption{\textbf{Prior optimizations for joint 3D/pose learning.}\vspace{-.5em}}
    \label{fig:opt_prior}
\end{figure}

\begin{figure}[t]
    \begin{subfigure}[t]{0.4\textwidth}
      \includegraphics[width=\columnwidth]{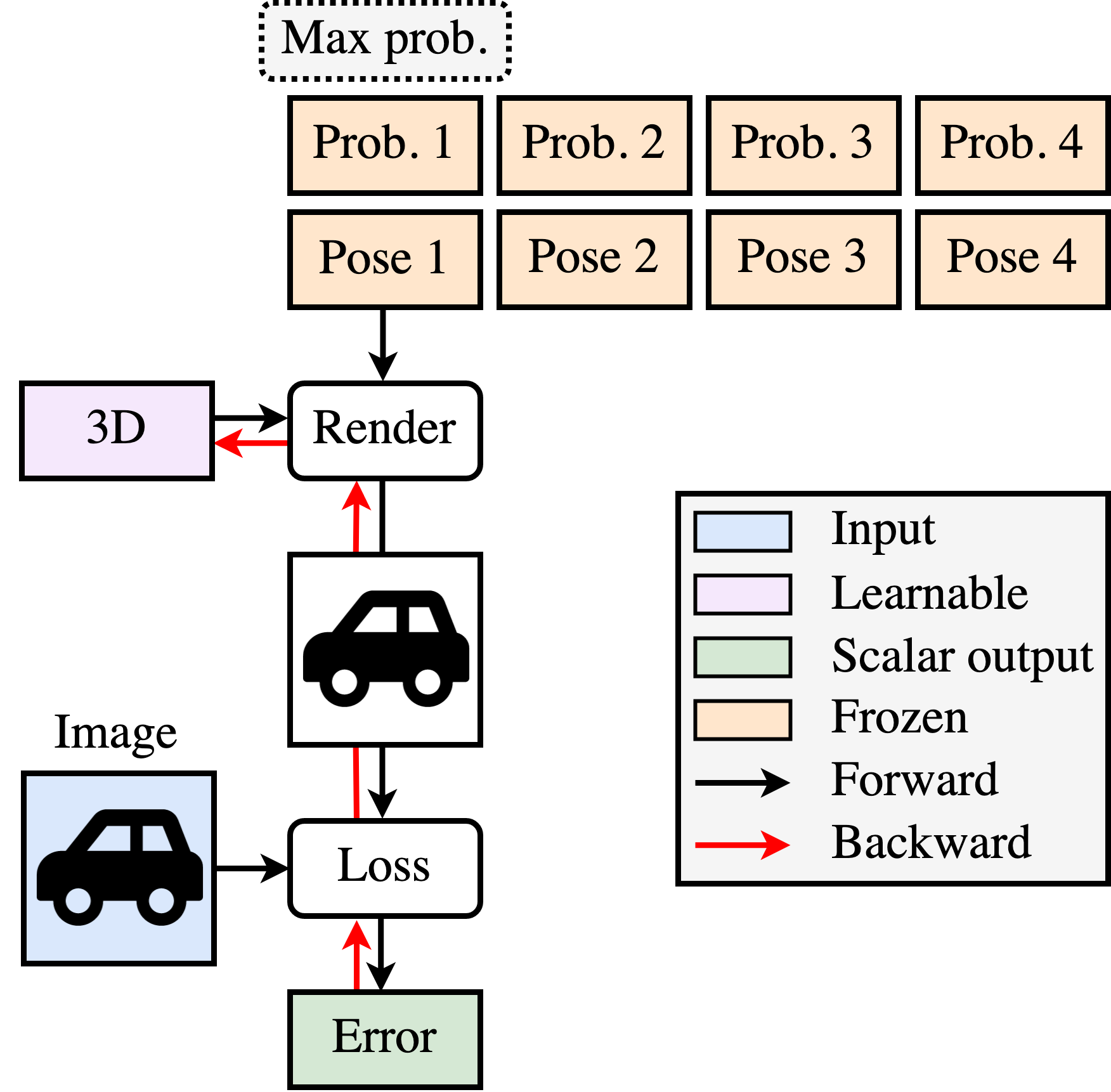}
      \caption{3D-step}
      \label{fig:opt_our1}
    \end{subfigure}\quad\quad\quad
    \begin{subfigure}[t]{0.495\textwidth}
      \centering
      \includegraphics[width=\columnwidth]{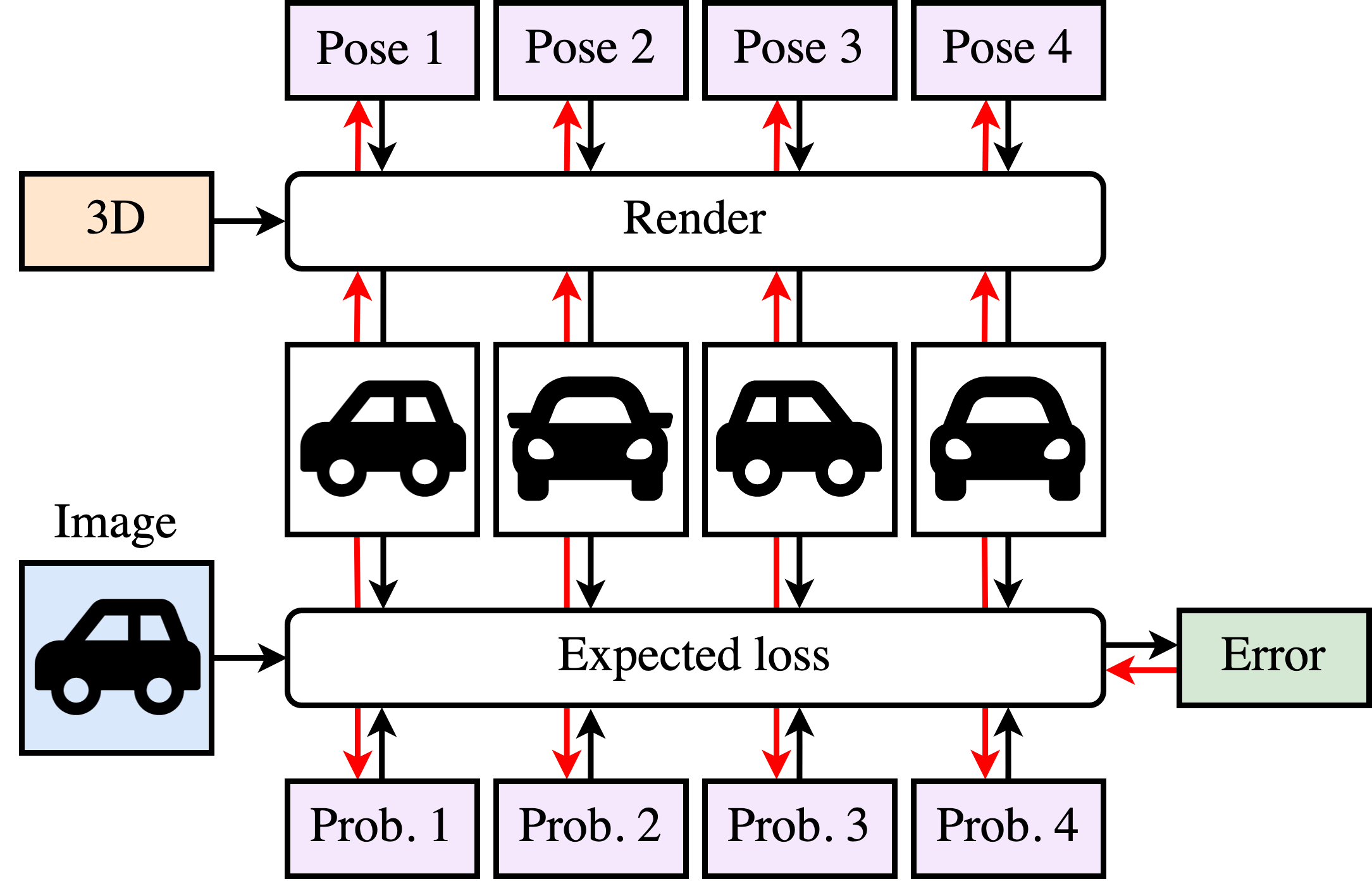}
      \caption{P-step}
    \label{fig:opt_our2}
    \end{subfigure}

    \caption{\textbf{Our alternate 3D / pose optimization.} Compared to prior works, we 
    propose an optimization that alternates between 2 steps. \textbf{(a)} We update the 3D 
  using the most likely pose candidate (3D-step). \textbf{(b)} We update the pose candidates 
and associated probabilities using the expected loss (P-step).\vspace{-1em}}
  \label{fig:opt_our}
\end{figure}

We analyze prior works on joint 3D and pose learning, illustrated in~\Cref{fig:opt_prior}, 
and compare them with our proposed optimization scheme, illustrated in~\Cref{fig:opt_our}.  
Prior optimization schemes can be split in two groups: (i) learning through the minimal error 
reconstruction~\cite{insafutdinovUnsupervisedLearningShape2018}, and (ii) learning through an 
expected error~\cite{hendersonLearningSingleimage3D2019, goelShapeViewpointKeypoints2020, 
tulsianiImplicitMeshReconstruction2020}.
  
In~\cite{insafutdinovUnsupervisedLearningShape2018}, all reconstructions associated to the 
different pose candidates are computed and both 3D and poses are updated using the 
reconstruction yielding the minimal error (see~\Cref{fig:opt_prior1}). We identified two 
major issues.  First, because the other poses are not updated for a given input, we observed 
that a typical failure case
corresponds to a collapse mode where only a single pose (or a small subset of poses) is used 
for all inputs.  Indeed, there is no particular constraint that encourages the use of all 
pose candidates. Second, inference is not efficient as the object has to be rendered from all 
poses to find the correct object pose.

In~\cite{hendersonLearningSingleimage3D2019, goelShapeViewpointKeypoints2020, 
tulsianiImplicitMeshReconstruction2020}, 3D and poses are updated using an expected 
reconstruction loss (see~\Cref{fig:opt_prior2}). While this allows to constrain the use of 
all pose candidates with a regularization on the predicted probabilities, we identified one 
major weakness common to these frameworks. Because the 3D receives gradients from all views, 
we observed a typical failure case where the 3D tries to fit the target input from all pose 
candidates yielding inaccurate texture and geometry. We argue such behaviour was not observed 
in previous works as they typically use a symmetry prior which prevents it from happening.  
Note that CMR~\cite{goelShapeViewpointKeypoints2020} proposes to directly optimize for each 
training image a set of parameters corresponding to the pose candidates. This procedure not 
only involves memory issues as the number of parameters scales linearly with the number of 
training images, but also inference problems for new images. To mitigate the issue, they 
propose to use the learned poses as ground-truth to train an additional network that performs 
pose estimation given a new image.

In contrast, our proposed alternate optimization, illustrated in~\Cref{fig:opt_our},  
leverages the best of both worlds: (i) 3D receives gradients from the most likely 
reconstruction, and (ii) all poses are updated using an expected loss. In practice, we 
alternate the optimization every new batch of inputs, and we define one iteration as either a 
a 3D-step or a P-step.

\section{Quantitative evaluation}\label{sec:eval}

\subsection{ICP alignment for better 3D evaluation}

In the main paper, we align shapes using our gradient-based version of the Iterative Closest 
Point (ICP)~\cite{beslMethodRegistration3D1992} with anisotropic scaling before evaluating 3D 
reconstructions. For consistency, we use the same protocol across benchmarks and advocate to 
do so for future comparisons. First, meshes are centered and normalized so that they exactly 
fit inside the cube of unit length $[-0.5, 0.5]^3$; this is important to obtain results that 
are comparable. Second, we sample 100k points on the mesh surfaces. Third, we run our ICP 
implementation which minimizes by gradient descent the Chamfer-$L_2$ distance between the 
point clouds by jointly optimizing 3 translation parameters, 6 rotation 
parameters~\cite{zhouContinuityRotationRepresentations2020} and 3 scaling parameters. In 
practice, we use Adam optimizer~\cite{kingmaAdamMethodStochastic2015}, a learning rate of 
0.01 and 100 iterations. Note that we use this gradient-based version of ICP instead of the 
classical iterative formulation as we found it to diverge when optimizing scale.

\begin{figure}[t]
    \centering
    \includegraphics[width=0.9\columnwidth]{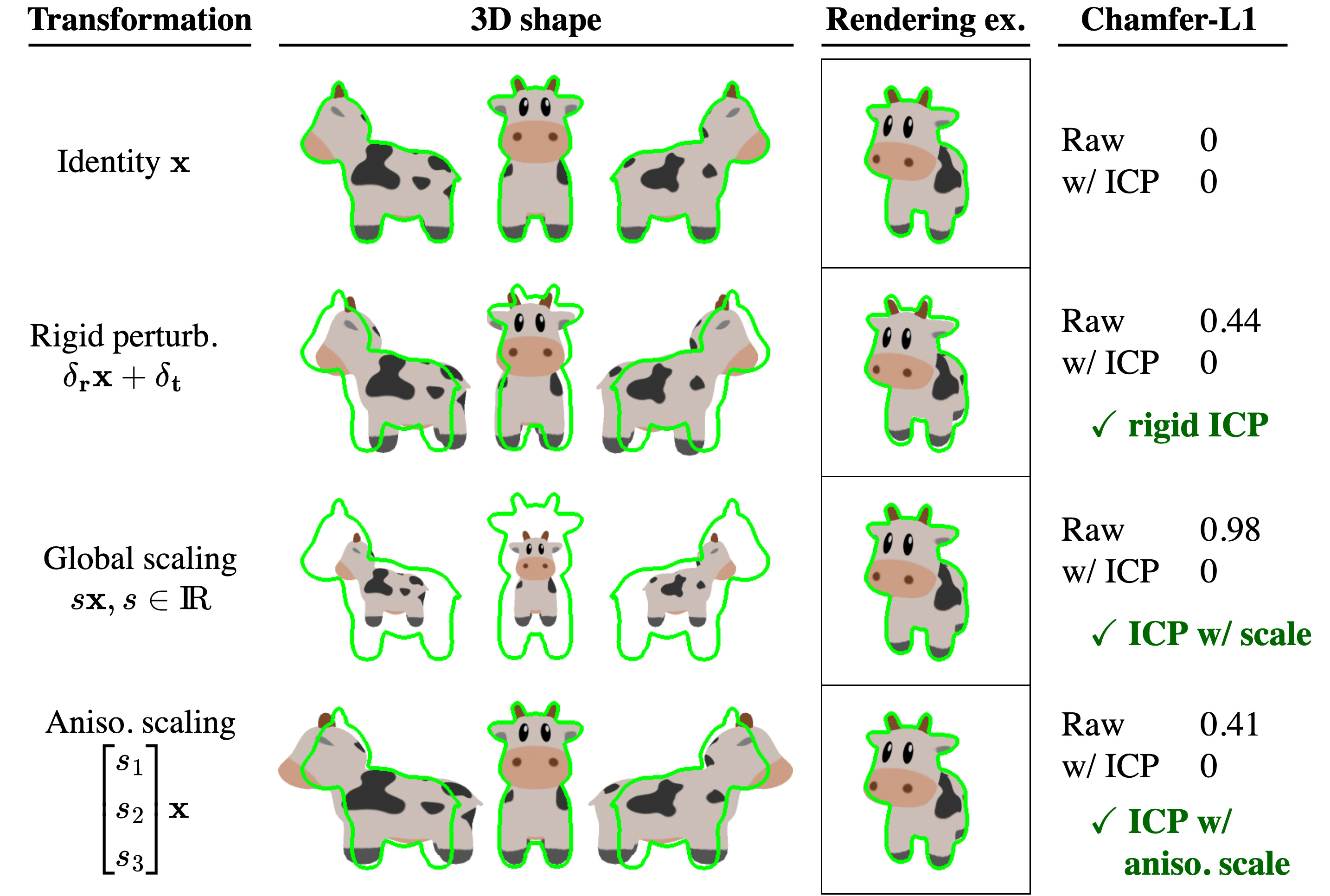}
    \caption{\textbf{3D reconstruction evaluation with and without ICP.} Rows correspond to 
    results obtained for transformed versions of a canonical shape and columns correspond to, 
  from left to right, the transformation used, resulting 3D shape, a rendering example and 
Chamfer-$L_1$ distance to the canonical shape. Green contours represent the shape and 
rendering from the canonical object for visual comparisons.}
    \label{fig:icp_eval}
\end{figure}

We argue that an ICP pre-processing is crucial for an unbiased 3D reconstruction evaluation 
and provide real examples in~\Cref{fig:icp_eval} to support our claim. Rows correspond to 
different transformations of the same canonical shape, and for each row, we show: the 
transformation used, the resulting 3D shape, a rendering example as well as Chamfer-$L_1$ 
distance to the canonical shape. We overlay the visuals with green contours representing the 
canonical shape and the canonical rendering for easier comparisons. We can make two important 
observations.  First, although all the transformed shapes are excellent 3D reconstructions of 
the canonical shape, they result in dramatically poor performances. As a comparison, these 
performances are similar to our ShapeNet results with ICP when the model outputs degenerate 
reconstructions.  Pre-processing the shapes using an ICP with anisotropic scaling mitigates 
this issue. Second, as shown by the rendering examples, for all these different shapes, we 
can find a pose that yields almost identical renderings. This hence emphasizes the numerous 
shape/pose ambiguities that arise from a given rendered image. As a result, it is extremely 
unlikely that a fully unsupervised SVR system predicting from a single image both the 3D 
shape and the pose will recover the exact pair of shape/pose used for annotations.  In this 
case, the cameras used for rendering are the same and we do not even consider focal 
variations, which raises even more ambiguities.

\subsection{ShapeNet results without ICP}

For completeness, we provide quantitative results obtained without ICP on the ShapeNet 
benchmark in~\Cref{tab:shapenet_noicp}. We indicate the supervision used and visually 
separate methods using multi-view supervision. In addition to methods compared in the main 
paper, we report (i) results from category-agnostic versions (Cat. agn) of 
DVR~\cite{niemeyerDifferentiableVolumetricRendering2020} and 
SoftRas~\cite{liuSoftRasterizerDifferentiable2019} presented 
in~\cite{niemeyerDifferentiableVolumetricRendering2020} and (ii) divergence results obtained 
by removing silhouette supervision from DVR and SoftRas.

\begin{table*}
  \centering
  \scriptsize
  \addtolength{\tabcolsep}{2pt}
  \begin{tabular}{@{}l|ccc|ccccc@{}}
  \toprule
  Method & \bf Ours & SDF-SRN & DVR & DVR & SoftRas & DVR & DVR & SoftRas\\

  Cat. agn. & & & & & & &\cmark & \cmark\\
  \mv & & & & \cmark & \cmark & \cmark & \cmark & \cmark \\
  \cam & &\cmark & \cmark & \cmark & \cmark & \cmark & \cmark & \cmark \\
  \sil & & \cmark & \cmark & & & \cmark & \cmark & \cmark \\

  \midrule
  airplane & 0.186 & 0.173 &\bf 0.157 & Div. & Div. & 0.151 & 0.190 &\bf 0.149 \\
  bench &\bf 0.257 & - & 0.386 & Div. & Div. & 0.232 &\bf 0.210 & 0.241 \\
  cabinet &\bf 0.284 & - & 0.849 & Div. & Div. & 0.257 &\bf 0.220 & 0.231\\
  car & 0.251 &\bf 0.177 & 0.282 & Div. & Div. & 0.198 &\bf 0.196 & 0.221 \\
  chair & 0.543 &\bf 0.333 & 0.464 & Div. & Div. & \bf 0.249 & 0.264 & 0.338  \\
  display &\bf 0.344 &  - & 0.968 & Div. & Div. & 0.281 &\bf 0.255 & 0.284 \\
  lamp & 0.987 & - &\bf 0.688 & Div. & Div. & 0.386 & 0.413 &\bf 0.381 \\
  phone &\bf 0.456 & - & 1.412 & Div. & Div. & 0.147 & 0.148 &\bf 0.131\\
  rifle &\bf 0.504 & - & 0.528 & Div. & Div. & \bf 0.131 & 0.175 & 0.155\\
  sofa &\bf 0.335 & - & 0.665 & Div.& Div. & \bf 0.218 & 0.224 & 0.407\\
  speaker &\bf 0.356 & - & 0.535 & Div. & Div. & 0.321 &\bf 0.289 & 0.320\\
  table &\bf 0.351 & - & 0.442 & Div. & Div. & 0.283 &\bf 0.280 & 0.374 \\
  vessel &\bf 0.384 & - & 0.400 & Div. & Div. & \bf 0.220 & 0.245 & 0.233\\

  \midrule

  mean &\bf 0.403 & - & 0.598 & Div. & Div. & \bf 0.236 & 0.239 & 0.266 \\
  \bottomrule
  \end{tabular}
  \caption{\textbf{ShapeNet comparison without ICP.} We report Chamfer-$L_1\downarrow$, 
    supervisions are: \multiview, \camerakeypoint, \silhouette. We separate methods using 
    multi-views  and \textbf{best} results are highlighted in each group.\vspace{-1.5em}
}

  \label{tab:shapenet_noicp}
\end{table*}

\section{Implementation details}\label{sec:implem}

\subsection{Modeling}

\subsubsection{Network architecture.} We use the same neural network architecture for all 
experiments. The encoder is composed of 4 CNN backbones followed by separate Multi-Layer 
Perceptron (MLP) heads predicting a rendering parameter. More specifically, the 4 backbones 
are respectively used to predict: (i) shape code $\betadef$ and scale $\scale$; (ii) texture 
code $\genbeta$; (iii) background code $\bkgbeta$; (iv) rotations $\rot_{1:K}$, translations 
$\trans_{1:K}$, and pose probabilities $\probpose_{1:K}$. Note that using a shared backbone 
instead of separated ones also yields great results and is advocated for decreasing the 
memory footprint and training time; the major benefit from using separated backbones is to 
produce
slightly more detailed textures and background. We follow prior works in 
SVR~\cite{groueixAtlasNetPapierMAch2018, meschederOccupancyNetworksLearning2019, 
niemeyerDifferentiableVolumetricRendering2020, goelShapeViewpointKeypoints2020} and use 
randomly initialized ResNet-18~\cite{heDeepResidualLearning2016} as backbone. Each MLP head 
has the same architecture with 3 hidden layers of 128 units and ReLU activations. The last 
layer of the MLP heads for shape, texture and background codes is initialized to zero to 
avoid discontinuity when increasing the size of the latent codes. The final activation of the 
MLP heads for scale, rotation, and translation is a $\tanh$ function and the output is scaled 
and shifted using predefined constants in order to control their range (see~\Cref{tab:design} 
for selected ranges). The learnable parts of the decoder are the shape deformation network 
$\dmlp$ and the two CNN generators $\tcnn$ and $\bcnn$ which respectively output $64 \times 
64$ images for texture and background. The MLP modeling the deformations has 3 hidden layers, 
ReLU activations and 128 units for real images; we use an increased number of units
for ShapeNet (512) which provides a small boost in performances. The CNN generators share the 
same architecture which is identical to the generator used in 
GIRAFFE~\cite{niemeyerGIRAFFERepresentingScenes2021}.  We refer the reader 
to~\cite{niemeyerGIRAFFERepresentingScenes2021} for details.

\vspace{-1em}
\subsubsection{Other design choices.} In all experiments, the predefined anisotropic scaling 
used to deform the icosphere into an ellipsoid is $[1, 0.7, 0.7]$. In~\Cref{tab:design}, we 
detail other design choices that are specific to all categories of  
ShapeNet~\cite{changShapeNetInformationRich3D2015} (second column) or all real-image datasets 
(third column). This notably includes a predetermined global scaling of the ellipsoid, a 
camera defined by a focal length $f$ or a field of view (fov), as well as scaling, 
translation and rotation ranges.

\begin{table*}[t]
  \centering
  \addtolength{\tabcolsep}{2pt}
  \begin{tabular}{@{}lcc@{}}
  \toprule
  Design type & ShapeNet & Real-image\\
  \midrule
  ellipsoid scale & 0.4 & 0.6\\
  camera & $f = 3.732$ & $\textrm{fov} = 30^\circ$\\
  $\scale_x / \scale_y / \scale_z$ & $1 \pm 0.5$ & $1 \pm 0.3$\\
  $\trans_x / \trans_y$ & $0 \pm 0.5$ & $0 \pm 0.3$\\
  $\trans_z$ (depth) & $2.732$ & $2.732 \pm 0.3$ \\
  $\rot_a$ (azimuth) & $[0^\circ, 360^\circ]$ & $[0^\circ, 360^\circ] $\\
  $\rot_e$ (elevation) & $30^\circ$ & $[-10^\circ, 30^\circ]$\\
  $\rot_r$ (roll) & $0^\circ$ & $[-30^\circ, 30^\circ]$\\

  \bottomrule
  \end{tabular}
  \caption{\textbf{Design choices.} Following standard 
    practices~\cite{liuSoftRasterizerDifferentiable2019, 
    niemeyerDifferentiableVolumetricRendering2020} on 
    ShapeNet~\cite{changShapeNetInformationRich3D2015}, we keep the default rendering values 
    used to generate the images for the focal length $f$, the distance to the camera 
    $\trans_z$ and the elevation $\rot_e$.  For real images, we keep the classical value of 
    2.732 for the distance to the camera $\trans_z$ and use a field of view (fov) of 
  $30^{\circ}$.   Note that we did not finetune these parameters, they were selected once 
through visual comparisons on a toy example. \vspace{-.5em}}
  \label{tab:design}
\end{table*}

\subsection{Training} 

In all experiments, we use a batch size of 32 images of size $64 \times 64$ and Adam 
optimizer~\cite{kingmaAdamMethodStochastic2015} with a constant learning rate of $10^{-4}$ 
that is divided by 5 at the very end of the training for a few epochs. The training 
corresponds to 4 stages where latent code dimensions are increased at the beginning of each 
stage and the network is then trained until convergence.  We use dimensions 0/2/8/64 for the 
shape code, 2/8/64/512 for the texture code, and 4/8/64/256 for the background code if any.  
In line with the curriculum modeling 
of~\cite{monnierDeepTransformationInvariantClustering2020}, we found it beneficial for the 
first stage to gradually increase the model complexity: we first learn to position the fixed 
ellipsoid in the image, then we allow the ellipsoid to be deformed, and finally we allow 
scale variabilities. In particular, we found this procedure prevents the model to learn 
prototypical shapes with unrealistic proportions. In the following, we describe other 
training details specific to ShapeNet~\cite{changShapeNetInformationRich3D2015} benchmark and 
real-image datasets.

\vspace{-1em}
\subsubsection{ShapeNet dataset.} We use the same training strategy for all categories. We 
train the first stage for 50k iterations, and each of the other stage for 250k iterations, 
where one iteration corresponds to either a 3D-step or a P-step of our alternate 
optimization.  We do not learn a background model as all images are rendered on top of a 
white background.  However, we found that our system learned in such synthetic setting was 
prone to a bad local minimum where the predicted textures have white regions that accommodate 
for wrong shape prediction.  Intuitively, this is expected as the system has no particular 
signal to distinguish white background regions from white object parts. To mitigate the 
issue, we constrain our texture model as follows: (i) during the first stage, the predicted 
texture image is averaged to yield a constant texture, and (ii) during the other stages, we 
occasionally use averaged textures instead of the real ones. More specifically, we sample a 
Bernoulli variable with probability $p = 0.2$ at each iteration and average the predicted 
texture image in case of success. We found this simple procedure to work well to resolve such 
shape/texture ambiguity.

\vspace{-1em}
\subsubsection{Real-image datasets.} We use the same training strategy for all real-image 
datasets. We train each stage for roughly 750k iterations, where one iteration either 
corresponds to a 3D-step or a P-step of our alternate optimization. Learning our structured 
autoencoder in such real-image scenario, without silhouette nor symmetry constraints, is very 
challenging.  We found our system sometimes falls into a bad local minimum where the texture 
model is specialized by viewpoints, \eg, dark cars always correspond to a frontal view and 
light cars always correspond to a back view. To alleviate the issue, we encourage uniform 
textures by occasionally using averaged textures instead of the real ones during rendering, 
as done on the ShapeNet benchmark. More specifically, we sample a Bernoulli variable with 
probability $p=0.2$ at each iteration and average the predicted texture image in case of 
success. We observed that it was very effective in practice, and we also found it helped 
preventing the object texture from modeling background regions. We do not use such technique 
in the last stage to increase the texture accuracy.

\end{document}